# A Comprehensive Review of Fish Feeding Behavior Analysis in Aquaculture: Tasks, Techniques, and Applications

Shulong Zhang[a, f], Daoliang Li[a, b, c, d, e, f], Jiayin Zhao[a, f], Mingyuan Yao[a, f], Yingyi Chen[a, b, c, d, e, f], Haihua Wang[a, b, c, d, e, f *]

*a National Innovation Center for Digital Fishery, Beijing 100083, China*

*b Key Laboratory of Smart Farming Technologies for Aquatic Animal and Livestock, Ministry of Agriculture and Rural Affairs, Beijing 100083, China*

*c State Key Laboratory of Efficient Utilization of Agricultural Water Resources, Beijing 100083, China*

*d Beijing Engineering and Technology Research Center for Internet of Things in Agriculture, Beijing 100083, China*

*e Key Laboratory of Digital Fisheries of Shandong Province, Yantai 264670, China*

*f College of Information and Electrical Engineering, China Agricultural University, Beijing 100083, China*

* Corresponding author.

Email: zhangshulong@cau.edu.cn (Shulong Zhang); dliangl@cau.edu.cn (Daoliang Li); zhaojiayin@cau.edu.cn (Jiayin Zhao); yaomingyuan@cau.edu.cn (Mingyuan Yao); chenyingyi@cau.edu.cn (Yingyi Chen); wang_haihua@163.com (Haihua Wang)

**Abstract:** Fish feeding behavior analysis is a key foundation for intelligent feeding and precision aquaculture management, and plays an important role in improving feed utilization efficiency, reducing production costs, and mitigating environmental burden. Existing reviews mainly focus on specific technical modalities or related applications in smart aquaculture, which makes it difficult to present the overall development of fish feeding behavior analysis in a comprehensive manner. To address these issues, this paper provides a thematic review of fish feeding behavior analysis in aquaculture, and systematically examines its task definition, technical support, and application status. First, from the task perspective, two core subtasks of fish feeding behavior analysis are clearly distinguished, and relevant behavioral characteristics and evaluation metrics are summarized. Second, from the technical perspective, the development trajectories of computer vision, acoustics, sensors, and multimodal fusion technologies are examined, and their advantages, limitations, and applicable scenarios are analyzed. On this basis, the application value of fish feeding behavior analysis in intelligent feeding and aquaculture management is further summarized. Finally, this paper discusses the challenges in robust perception under complex environments, generalization across fish species and farming scenarios, collaborative multimodal modeling and lightweight deployment, closed loop intelligent feeding, coordinated optimization of multiple tasks, and long-term production validation, and outlines future research directions. This review provides a reference for task standardization, technical selection, and engineering application in fish feeding behavior analysis, and offers insights into the development of smart aquaculture and sustainable aquaculture management.

## 1. Introduction

Aquaculture plays a crucial role in satisfying the growing global demand for fish and providing a sustainable food source [1]. According to the Food and Agriculture Organization of the United Nations [2], the total global fishery and aquaculture production has reached 223.2 million tons, with aquaculture production surpassing wild capture fisheries for the first time. The per capita protein supply from aquatic animal foods is at least 20% of the per capita protein supply from all animal sources. However, with the expansion of farming scale and increased intensification, issues such as feed waste, rising disease risks, growing environmental burdens, and more complex production management have become increasingly prominent [3,4]. Among these, feeding management not only directly affects fish growth and farming profitability, but is also closely related to water quality stability, environmental emissions, and fish welfare. Therefore, accurately perceiving the feeding status of fish schools and enabling demand-based feeding have become key issues in precision aquaculture and sustainable aquaculture management.

Fish feeding behavior is an important external representation of feeding demand, metabolic state, environmental adaptability, and health status in fish school. Changes in swimming speed, spatial distribution, group aggregation, feeding response, acoustic features, and physiological parameters during feeding can reflect feeding motivation and physiological state [5,6]. Therefore, continuous, objective, and quantitative analysis of fish feeding behavior helps identify real time feeding demand, optimize feeding timing and ration size, reduce feed waste, lower the environmental burden of aquaculture, and improve production efficiency and management precision.

Early assessment of fish feeding behavior and feed delivery mainly relied on manual observation and experience-based judgment. Farmers usually estimate the feeding status of fish schools according to external phenomena such as feeding frequency, feeding duration, surface disturbance, and group activity. However, this approach is highly subjective, is easily affected by observer experience, environmental conditions, and farming scale, and cannot meet the requirements of long term, continuous, and high precision monitoring. With the development of automated equipment, feeding practices gradually shifted from manual feeding to timed and rationed mechanical feeding [7]. Although mechanical feeding can reduce labor costs, it remains essentially rule driven and is difficult to adjust dynamically according to real time fish behavior and feeding demand, which may lead to underfeeding or overfeeding. In this context, fish feeding behavior analysis based on intelligent perception and data driven modeling has become an important research direction in precision aquaculture. In recent years, computer vision [8–12], acoustics [13–16], sensors [17–19], and multimodal fusion methods [20–23] have been widely used for feeding state recognition, feeding intensity estimation, and intelligent feeding decision making, providing new technical routes for automated monitoring of fish feeding behavior and optimization of aquaculture management.

Despite rapid progress in related research, existing reviews still fail to present a comprehensive picture of the overall development of fish feeding behavior analysis. Most reviews often focus on single-modality or generalized digital aquaculture tasks. For example, An et al. [24] and Zhang et al. [25] systematically summarized the applications of computer vision technology in fish target detection, body length and weight estimation, feeding behavior analysis, and intelligent feeding. Li et al. [26] discussed the application of acoustic technology in fish biomass estimation, behavior analysis, and welfare improvement. Li et al. [27] analyzed the applications of single-modal and multimodal fusion technologies in water quality monitoring, feeding behavior analysis, disease prediction, and biomass estimation. Cui et al. [28] considered fish tracking, counting, and behavior analysis as related tasks in digital aquaculture, reviewing methods such as vision, acoustics, and biosensors. These studies provide an important basis for understanding the development of smart aquaculture technologies. However, fish feeding behavior analysis is usually treated as only one subtask, and the related discussion remains scattered. Although Li et al. [29] provide an early systematic review of computer vision, acoustic techniques, and sensors in fish feeding behavior research, the technical scope and frontier coverage of that review no longer fully reflect the current state of the field because of rapid methodological advances. Recent reviews focus mainly on computer vision methods and have not incorporated acoustics, sensors, and multimodal fusion into a unified analytical framework [30]. Therefore, a focused review that takes fish feeding behavior analysis as the central object and combines task systematicity, technical completeness, and application orientation is still lacking.

In addition, existing studies often discuss feeding behavior recognition and feeding intensity quantification together, without clearly distinguishing their task boundaries, progressive relationship, and methodological differences. In fact, behavior recognition and feeding intensity quantification are two core subtasks in fish feeding behavior analysis that are closely related but differ in their objectives. Behavior recognition mainly focuses on whether feeding behavior occurs and which feeding related state the fish are in, and its output is usually a set of discrete categories. Feeding intensity quantification further focuses on the activity level, duration, group response strength, and dynamic change trend of feeding activity, and its output can be expressed as intensity levels, continuous scores, or feeding decision indicators. These two subtasks differ substantially in task objectives, data features, model design, and evaluation metrics. Without a clear distinction, it is difficult to understand the research trajectory of this field in a systematic manner, and the rational application of different technical routes in intelligent feeding and aquaculture management may also be limited.

Based on the above context, this paper provides a focused review of fish feeding behavior analysis in aquaculture, aiming to build an integrated framework that connects task definition, technical support, and management application. The main contributions of this paper are reflected in three aspects. First, from the task perspective, this paper systematically defines the core subtasks of fish feeding behavior analysis, clarifies the relationship and differences between behavior recognition and feeding intensity quantification, and summarizes related behavioral representation indicators and evaluation metrics. Second, from the technical perspective, this paper reviews the development of computer vision, acoustic techniques, sensor technologies, and multimodal fusion methods, and compares the advantages and limitations of different technical routes in data acquisition, feature representation, model construction, and application scenarios. Finally, from the application perspective, this paper summarizes the value of fish feeding behavior analysis in intelligent feeding and aquaculture management, and discusses the key challenges and future research directions, with the aim of providing a systematic reference for research on fish feeding behavior analysis and smart aquaculture.

The remainder of this paper is organized as follows. Section 2 defines the core tasks of fish feeding behavior analysis and summarizes behavioral representation indicators and evaluation metrics. Section 3 systematically reviews the development of computer vision, acoustic techniques, sensor technologies, and multimodal fusion methods. Section 4 summarizes their applications in intelligent feeding and aquaculture management. Section 5 discusses current challenges and future research directions. Section 6 concludes this paper.

## 2. Task definition of fish feeding behavior analysis

### 2.1 Definition of fish feeding behavior analysis

Fish feeding behavior refers to the individual or group activity patterns that fish exhibit when they perceive, approach, capture, and ingest feed. These patterns include feeding actions, feeding frequency, feeding duration, spatial aggregation, swimming responses, and interactions among individuals or between fish schools and the environment. Compared with general fish behavior analysis, feeding behavior analysis places greater emphasis on the identification and interpretation of feeding related states, feeding processes, and feeding demand. Its core objective is to use behavioral changes, physiological responses, and environmental disturbance signals generated by during the feeding to support objective perception, quantitative characterization, and decision making related to feeding management. Compared with traditional feeding judgment methods that rely on manual experience, fish feeding behavior analysis emphasizes continuous monitoring, quantitative assessment, and dynamic feedback based on objective data. Therefore, fish feeding behavior analysis is not only an important research topic in fish ethology, but also an important technical foundation for intelligent feeding, fish school health assessment, and precision aquaculture management.

### 2.2 Task forms of fish feeding behavior analysis

From the perspective of task composition, fish feeding behavior analysis mainly includes two core subtasks, namely feeding behavior recognition and feeding intensity quantification, as shown in Figure 1. Feeding behavior recognition mainly focuses on determining whether feeding behavior occurs, identifying which behavioral state the fish are in, or further distinguishing different types of feeding-related states [31–33]. This task is essentially a behavioral state discrimination problem, and its outputs are usually discrete category labels. Common task forms include feeding/non-feeding binary classification and multi-class behavior recognition. Feeding/non-feeding binary classification only determines whether feeding behavior occurs, and is suitable for coarse-grained discrimination of feeding activity. Therefore, it is often used in automatic feeding control and basic state monitoring [34–36]. Multi-class behavior recognition further extends the representation of behavioral states, and usually divides fish feeding behavior into multiple state categories, such as pre-feeding, feeding start, feeding, and post-feeding [37,38,32], thereby providing a more detailed description of the stage-wise changes in fish school feeding behavior.

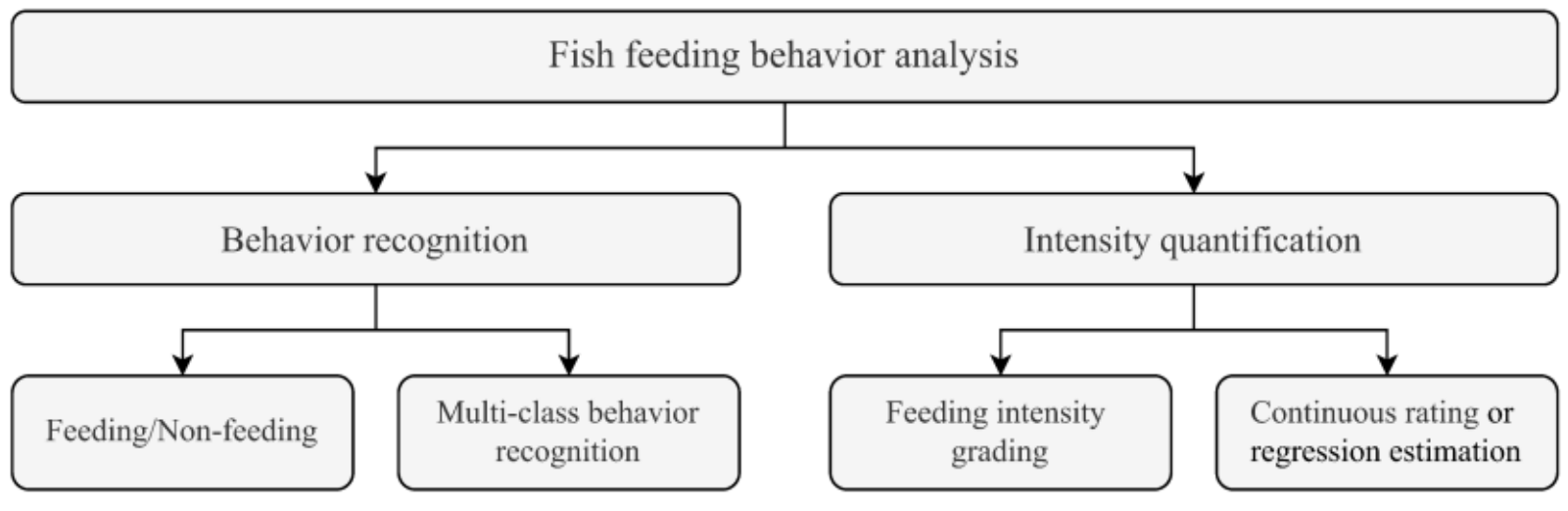

Figure 1. Task forms of fish feeding behavior analysis.

Feeding intensity quantification not only concerns whether fish are feeding, but also focuses on estimating the current feeding activity of fish schools, the changing trend of feeding demand, and the corresponding need for feeding regulation. Common task forms include intensity level classification, continuous scoring, and regression estimation. Intensity level classification usually divides feeding intensity into several discrete levels, such as none, weak, medium, strong, according to group activity, feeding response level, or feature changes, thereby supporting ration adjustment and precision management [39,9,40,11,41]. Continuous scoring or regression estimation represents fish feeding intensity, feeding willingness, or feeding demand in the form of continuous values. By establishing mappings between behavioral features, multisource perception signals, and continuous output variables, these methods enable fine-grained quantification of feeding status [42,43,38,44,45]. For example, Liu et al. [46] divided fish feeding intensity into 10 score levels. Compared with discrete classification tasks, continuous scoring provides higher-resolution analytical results and therefore has greater application value in precision feeding decision support.

Although the two tasks differ in terms of objectives and output forms, they are closely related in analytical logic. Feeding behavior recognition provides the basis for feeding intensity quantification, since the feeding degree and changing trend can be further evaluated only when fish schools are identified as being in feeding related states. Feeding intensity quantification extends the analytical objective from static judgment about whether feeding occurs to more application-relevant questions, including how strong the feeding activity is, how feeding demand changes, and whether the feeding strategy should be adjusted. Therefore, these two subtasks are not independent of each other, but jointly constitute the basic task framework of fish feeding behavior analysis.

## 2.3 Representation indicators of fish feeding behavior analysis

Fish feeding behavior analysis relies on multiple types of representation indicators, which reflect feeding demand, feeding willingness, feeding status, and the interactions between fish behavior and the farming environment from different perspectives [30,47,48]. According to differences in indicator sources and represented objects, the feeding behavior representation indicators used in existing studies can be broadly grouped into three categories, namely physical representations, physiological representations, and environmental representations, as shown in Figure 2.

Physical representations are the most commonly used type in current research. They primarily describe the external movement and spatial distribution characteristics of individual fish or fish schools during the feeding process, including speed [49,50], acceleration [51], group activity level [52], aggregation degree [53–55], direction and angle changes [56], and spatial distribution [57,58]. These indicators can intuitively reflect the response strength and feeding willingness of fish schools to feeding stimuli, and are therefore widely used in fish feeding behavior analysis.

Physiological representations mainly reflect changes in the internal state, feeding action characteristics, and acoustic responses of fish during feeding, including heart rate [59], tail beat frequency [60], mandibular vibration frequency [61], sound [16,62,63], and body surface color changes [64]. Compared with physical representations, these indicators reveal fish responses to feed stimuli from perspectives that are more closely related to individual physiological states or feeding action mechanisms. Therefore, they have distinct advantages in assessing feeding motivation, distinguishing feeding behavior states, and analyzing appetite changes.

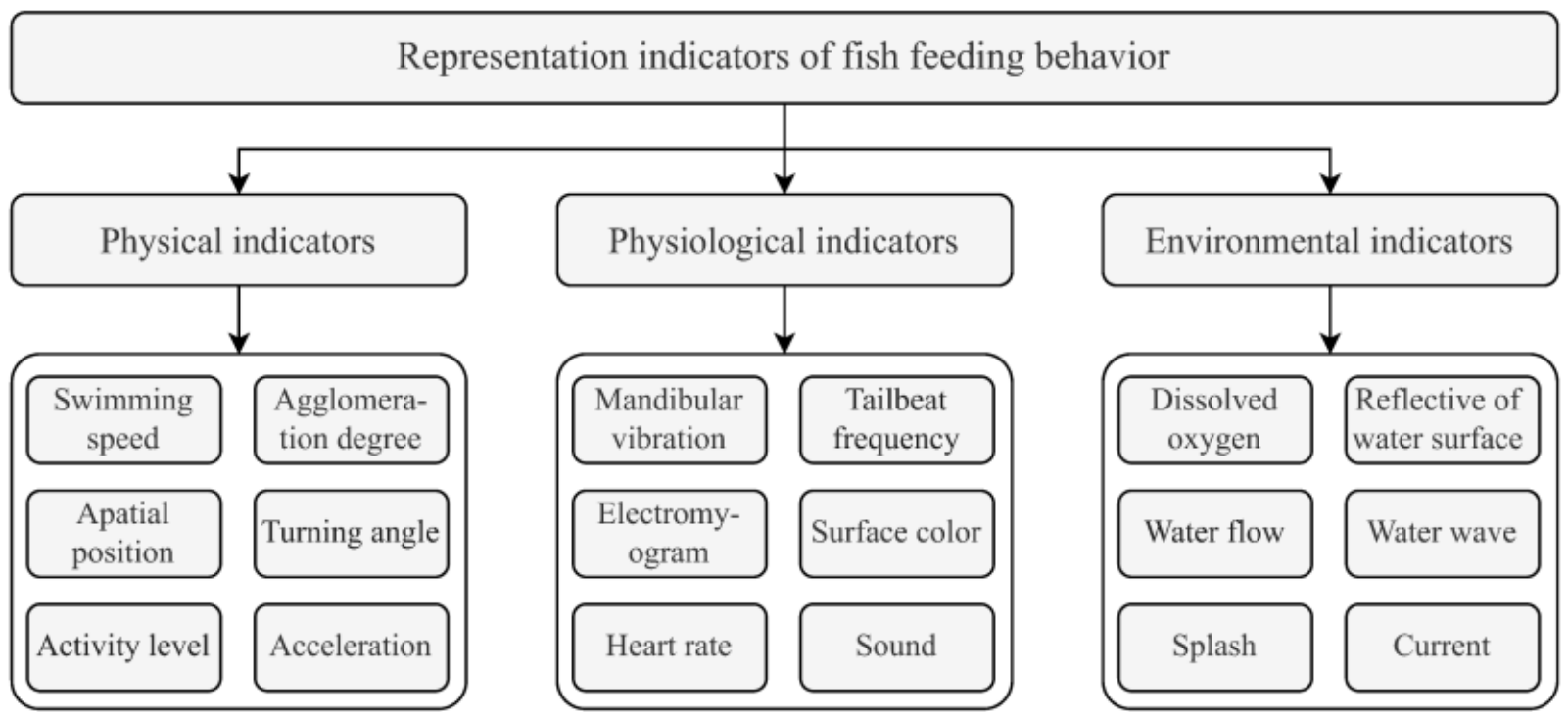

Figure 2. Representation indicators of fish feeding behavior.

Environmental representations mainly describe changes in the water body, water surface, or farming system caused by fish feeding activity, including dissolved oxygen [65,66], water flow [67], water splashes [42,68,69], water waves [70,71,18,72], water surface reflection [73], and current [17]. Although these indicators do not directly originate from the fish body itself, they can indirectly reflect the occurrence, intensity, and duration of feeding behavior through changes in external media. Compared with direct observation of fish behavior, environmental representations offer certain advantages in scenarios with turbid water, severe group occlusion, difficult individual separation, or limited imaging conditions. Therefore, they have considerable application potential in practical aquaculture environments.

## 2.4 Evaluation metrics of fish feeding behavior analysis

In correspondence with behavioral representation indicators, evaluation metrics for fish feeding behavior analysis usually depend on the task type, output form, and application objective. For classification tasks such as feeding/non-feeding discrimination, multi-class behavior recognition, and intensity level classification, commonly used metrics include accuracy, precision, recall, and F1 score. These metrics are used to measure the overall performance of models in behavioral category discrimination and their ability to distinguish different classes. For tasks such as continuous scoring or regression estimation of feeding intensity, mean absolute error, Root Mean Squared Error (RMSE), and the coefficient of determination are commonly used to evaluate the deviation and goodness of fit between predicted values and ground truth values. In addition, for visual tasks such as object detection, individual tracking, and feeding region recognition, some studies also adopt metrics such as intersection over union, mean Average Precision (mAP), tracking accuracy, and identity preservation ability to evaluate the reliability of the perception stage.

Since fish feeding behavior analysis ultimately serves intelligent feeding and online aquaculture management, algorithmic performance metrics alone are insufficient for fully assessing its practical value. Therefore, practical applications should also consider real time and engineering metrics, including model inference speed, processing latency, frame rate, system response time, edge deployment complexity, and long-term operational stability. From the perspective of aquaculture management, production performance indicators should also be further considered, including residual feed, feed utilization efficiency, Feed Conversion Ratio (FCR), growth performance, water quality changes, and feeding cost. Only by combining algorithm evaluation, system evaluation, and farming outcome evaluation can the application feasibility and promotion value of fish feeding behavior analysis techniques in real aquaculture scenarios be assessed more comprehensively.

# 3. Technical support of fish feeding behavior analysis

## 3.1 Fish feeding behavior analysis based on computer vision

Computer vision is one of the most widely used and mature technical routes for fish feeding behavior analysis. Its core idea is to collect image and video data of fish schools during feeding through above-water or underwater cameras, and to extract visual information such as fish motion, group distribution, water surface disturbance, residual feed, and changes in feeding regions. These cues are then used to analyze feeding behavior states and feeding intensity, as shown in Figure 3. Compared with manual observation, computer vision offers several advantages, including non-contact sensing, rich information content, continuous recording, and easy integration with automated systems. Therefore, it has been widely used in research on intelligent feeding and precision aquaculture.

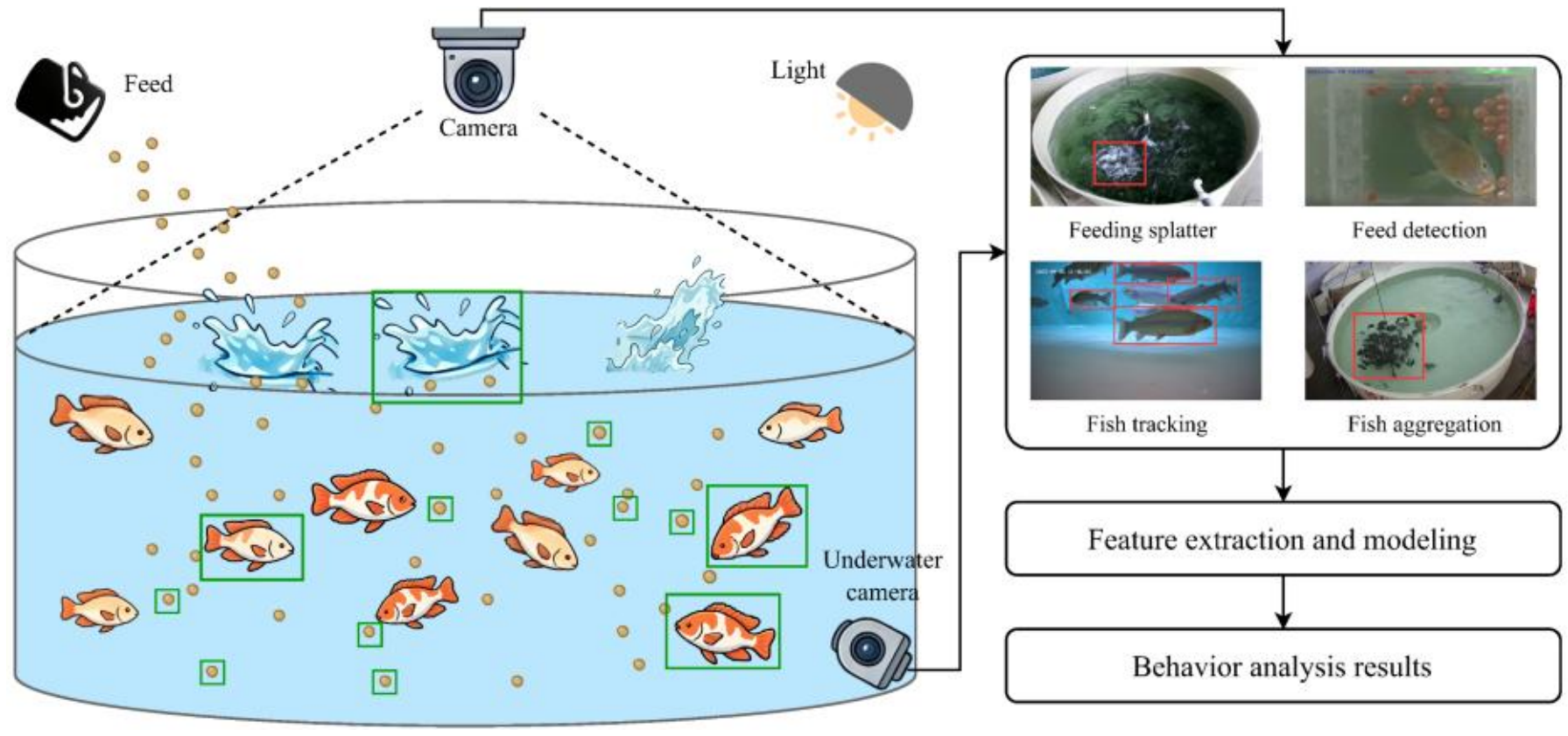

Figure 3. A framework for analyzing fish feeding behavior based on computer vision.

### 3.1.1 Fish feeding behavior recognition based on computer vision

Fish usually exhibit different group shapes, spatial distributions, swimming speeds, aggregation degrees, and water surface disturbance patterns at different feeding stages. Therefore, visual information provides an important basis for feeding state discrimination. Early studies mainly rely on predefined motion features and image texture features, and recognize feeding behavior through foreground segmentation, object extraction, and feature classification [74,56]. These methods have a certain degree of interpretability, but usually require clear fish contours or stable water surface images. When illumination changes, water surface fluctuations, reflection, bubbles, or occlusion are severe, segmentation errors further affect recognition stability and computational efficiency. To reduce the dependence on accurate fish segmentation, some studies use frequency domain analysis or feature matrix modeling to identify the fish feeding state [75,76]. These methods start from global image changes or group motion patterns, which reduces the difficulty of detecting and segmenting individual fish. However, their applicability still depends on scene stability and the quality of feature design.

With the development of deep learning, research attention gradually shifts toward data driven feature learning. Yang et al. [37] introduced spatial and channel attention mechanisms on top of EfficientNet-B2 to enhance the model's attention to the feeding area. Zhang et al. [77] combined MobileNetV2 and SENet to construct a recognition model of fish feeding behavior, which achieves high recognition accuracy, but still has certain limitations in inference speed. To balance recognition accuracy and real time deployment, Yang et al. [78] added identity and residual branches to RepVGG and introduced the Efficient Channel Attention (ECA) mechanism, balancing recognition accuracy and inference speed to some extent. In addition, some studies improved image segmentation and behavioral region modeling methods [34,79], which enhances the stability of visual inputs and provides a more reliable basis for subsequent feeding state recognition.

Since fish feeding behavior shows clear temporal continuity, relying only on single-frame images is often insufficient to fully describe the process of behavioral evolution. Therefore, temporal modeling of video data has gradually become an important direction in feeding behavior recognition. Måløy et al. [80] proposed a Dual-Stream Recurrent Network (DSRN) to recognize salmon feeding behavior, but its structure is relatively complex, which limits its applicability to long videos or resource constrained scenarios. To simplify the network structure and improve the utilization of spatiotemporal features, Zhang et al. [81] used Variational Auto Encoder (VAE) to extract Gaussian mean and variance vectors from video frames, construct a feature matrix, and then feed it into a CNN for recognition. Wang et al. [82] introduced efficient channel attention and residual modules to improve the Boundary Matching Network (BMN), thereby reducing detection errors during feeding state transitions. Zheng et al. [83] constructed a Spatiotemporal Attention Network (STAN), emphasizing that feeding behavior recognition should focus more on overall relationships within fish school rather than local image features. To reduce dependence on manually annotated data, Wang et al. [38] proposed an unsupervised recognition method based on an Appearance-motion Autoencoder Network (AMA-Net). In addition, Tan et al. [32] used You Only Look Once (YOLO) to accurately distinguish the hunger, feeding, and post-feeding stages of juvenile grouper. Table 1 summarizes representative works on fish feeding behavior recognition based on computer vision.

Table 1. Representative works on fish feeding behavior recognition based on computer vision.

| References | Fish species | Application scenarios | Dataset type | Behaviors | Models | Accuracy |
|---|---|---|---|---|---|---|
| [80] | *Salmo salar* | -- | 76 videos | Feeding, Non-feeding | DSRN | 80.00% |
| [81] | *Salmo salar* | Indoor factory farming | 3891 videos | Feeding, Non-feeding | VAE-CNN | 89.00% |
| [75] | *Oreochromis* | Indoor factory farming | 300 images | Feeding, Non-feeding | SVM | 99.24% |
| [76] | *Cyprinus carpio* | Laboratory | 1000 images | Stress, Feeding | PCA | 96.02% |
| [37] | *Oplegnathus punctatus* | Indoor factory farming | 15150 images | Feeding-start, Feeding-before, Feeding, Post feeding, Frightening | EfficientNet-B2 | 89.56% |
| [84] | *Oplegnathus punctatus* | Indoor factory farming | 10027 images | Normal feeding, Normal non-feeding, Stress non-feeding, Stress feeding | GhostNet | 98.12% |
| [77] | *Plectropomus leopardus* | Indoor factory farming | 3720 images | Feeding, Non-feeding | MobileNetV2-SENet | 97.76% |
| [38] | *Oplegnathus punctatus* | Indoor factory farming | 267 videos | Before, during and after feeding | AMA-Net | 99.71% |
| [82] | *Golden trout* | Laboratory | 100 videos | Feeding, Non-feeding | BMN-Fish | 91.11% |
| [83] | *Pompanos* | Aquaculture vessel | 23000 images | Feeding, Non-feeding | STAN | 97.97% |
| [34] | *Oplegnathus punctatus* | Indoor factory farming | 1400 images | Feeding, Non-feeding | ECA-Deeplabv3+ | 94.97% |
| [33] | -- | -- | 297029 images | No or weak feeding, Food picking, Swimming, Predatory | TMVF | 96.56% |
| [36] | *Largemouth Bass* | Indoor factory farming | 2160 images | Feeding, Non-feeding | SA-RepVGG-PLK | 95.80% |
| [32] | *Hybrid grouper* | Laboratory | 531 images | Hunger, Feeding, Post-feeding | YOLO + RF | 95.00% |
| [85] | *Crucian carp, grass carp* | Outdoor ponds | 1511 images | Feeding, Non-feeding | FFishNet-YOLOv8 | mAP=95.59% |
| [86] | *Micropterus salmoides* | Indoor factory farming | 5200 images | Feeding, Non-feeding | YOLOv8-FB | mAP=94.80% |

Note: -- indicates that the data cannot be obtained. SVM (Support Vector Machine); PCA (Principal Component Analysis); TMVF (Trusted Multi-View Fish); SA (Shuffle Attention); PLK (Pyramidal Lucas–Kanade); RF (Random Forest).

### 3.1.2 Fish feeding intensity quantification based on computer vision

After exposure to feeding stimuli, fish usually show accelerated swimming, rapid aggregation, stronger food competition, and intensified water surface disturbance. As satiety increases, group activity decreases, spatial distribution becomes more dispersed, and fish gradually return to routine swimming states. Therefore, visual features such as fish motion, group aggregation, and water surface response are often used to infer feeding intensity and appetite changes. Hu et al. [87] analyzed the aggregation degree of the fish school and the splash area generated during feeding, and used the area ratio between them to represent the hunger and satiety status of fish schools. Zhao et al. [88] quantified the fish appetite from the perspectives of dispersion degree, interaction force, and water flow field changes. Chen et al. [89] and Chen et al. [90] quantified feeding intensity using texture variation and weighted fusion of multiple features, respectively. To improve applicability in practical aquaculture scenarios, Zhao et al. [72] introduced a wave height correction factor to optimize the texture feature model, enabling effective quantification of feeding behavior in large yellow croaker under real sea conditions. These methods usually have strong interpretability and can directly associate visual features with feeding activity. However, they are sensitive to imaging conditions, threshold settings, and environmental stability, and still face limitations in complex water surfaces or cross scenario applications.

In addition to directly analyzing fish motion, some studies indirectly infer feeding intensity from auxiliary visual phenomena related to feeding. Zhao et al. [73] assessed the fish feeding intensity through changes in water surface reflection regions. Zheng et al. [91] inferred the satiation state according to the time required for feed to sink to a threshold depth, but this method is only applicable to fish that feed at the water surface and then move toward the bottom, such as carp and channel catfish. Hu et al. [92] used wave size caused by feeding to determine whether feeding should continue. Wu et al. [69] proposed a feeding intensity quantification method based on feeding splash thumbnails, which effectively reduced interference from water surface reflections, light spots, and ripples. However, in low-density farming environments or for fish fry, where splashing phenomena are not prominent, the model's performance may decline.

Deep learning models can automatically learn high level features related to feeding intensity from complex visual inputs, and can directly quantify fish feeding intensity. Zhou et al. [93] used a CNN based on the LeNet5 architecture to achieve automatic classification of fish feeding intensity. Xu et al. [94] improved the long convolutional recurrent network to enhance robustness under small sample scenarios. Kong et al. [39] combined active learning with VGG16 to reduce labeling costs. Zhang et al. [95] proposed MSIF-MobileNetV3, which is suitable for feeding scenarios where fish targets are small and swimming randomly. In addition, the introduction of attention mechanisms and Transformer structures further improves the perception of key regions and global feature representation under complex backgrounds [96,97,11].

For video sequence modeling, Ubina et al. [98] input optical flow frames into a 3DCNN to achieve high-precision quantification of feeding intensity. Zhao et al. [99] proposed an appetite assessment method based on individual behavior, which uses ByteTrack and Spatiotemporal Graph Convolutional Network (ST-GCN) for tracking and extracting the movement characteristics, effectively avoiding information loss caused by fish schools stacking. To reduce the number of parameters and computational resource consumption of deep learning models, lightweight models have been proposed successively, such as 3D ResNet-GloRe [100], 2D Motion-EfficientNetV2 [101], YOLOv8n [102], and YOLO-Feed [103]. In addition, Xu et al. [104] integrated the Convolutional Block Attention Module (CBAM) and Bi-directional Long Short-Term Memory (BiLSTM) into MobileViT to realize the collaboration of spatial feature extraction and temporal dependency modeling. Wu et al. [41] further proposed Spatiotemporal Cooperative Attention (STCA), which suppresses redundant frame noise and enhances attention to key feeding areas, further improving the accuracy and robustness of feeding intensity quantification in video sequences. Table 2 summarizes representative works in fish feeding intensity quantification based on computer vision.

**3.1.3 Summary**

Computer vision-based fish feeding behavior analysis focuses on capturing the spatial structure, motion patterns, and feeding related visual phenomena of fish schools, and its technical evolution is mainly reflected in feature representation, analytical targets, and model design. Early studies rely on features such as color, texture, aggregation index, splash area, and ripples. Although these features have strong interpretability, they are sensitive to imaging conditions and environmental changes. With the introduction of deep learning, visual methods can automatically learn higher level discriminative features, and combine 3DCNN, LSTM, optical flow, object tracking, spatiotemporal attention, and GCN to capture the dynamic evolution and state transitions of fish behavior. The use of lightweight networks, attention mechanisms, Transformer structures, and weakly supervised or unsupervised learning methods further promotes the development of computer vision techniques toward real time monitoring, low annotation cost, and edge deployment. However, the application of visual methods in real aquaculture environments still faces multiple challenges. Turbid water, suspended particles, bubbles, light reflection, and water surface fluctuations can significantly reduce image quality, thereby affecting feature extraction and model stability. Under high density farming conditions, fish occlusion and group overlap are severe, making individual level behavior extraction and stable tracking difficult. Different fish species, feeding modes, farming systems, and camera deployment conditions can cause clear domain shifts, which limits the cross-scenario generalization ability of models. Although many deep models perform well on experimental datasets, their real time inference on edge devices, long term stable operation, and ability to support explainable feeding decisions still require further validation.

Table 2. Representative works on fish feeding intensity quantification based on computer vision.

| References | Fish species | Application scenarios | Dataset type | Intensity level | Models | Accuracy |
|---|---|---|---|---|---|---|
| [105] | *Salmo salar* | Laboratory | 625 videos | None, Weak, Medium, Strong | MKEM-RNN | 98.31% |
| [106] | *Oplegnathus punctatus* | Indoor factory farming | 1075 images | Active, Not active | Multi-task neural network | 95.44% |
| [39] | *Oplegnathus punctatus* | Indoor factory farming | 3000 images | None, Weak, Medium, Strong | VGGAL-Mish-RAdam-Entselect | 99.60% |
| [100] | *Trout* | Laboratory | 400 videos | None, Weak, Medium, Strong | 3D ResNet-Glore | 92.68% |
| [43] | *Oplegnathus punctatus* | Indoor factory farming | 1415 images | Hunger, 30% full, 70% full, full, overeating | ResNet-CA | 93.40% |
| [107] | *Anguilla japonica* | Indoor factory farming, Outdoor ponds | 2000 images | Strong, Relatively strong, Normal, Relatively weak, and weak | Double-flow residual CNN | 98.60% |
| [95] | *Oplegnathus punctatus* | Indoor factory farming | 12000 images | None, Weak, Medium, Strong | MSIF-MobileNetV3 | 96.40% |
| [8] | *Oplegnathus punctatus* | Indoor factory farming | 1038 images | None, Weak, Medium, Strong | BlendMask-VoVNetV2 | 83.70% |
| [69] | *Micropterus salmoides* | Outdoor ponds | 31000 images | Low, Medium, High | Semi-supervised learning | 98.22% |
| [101] | *Micropterus salmoides* | Indoor factory farming | 743 videos | None, Weak, Strong | Motion-EfficientNetV2 | 93.97% |
| [108] | *Oreochromis* | Indoor factory farming | 16000 images | None, Weak, Medium, Strong | ResNet34-T-CBAM | 99.72% |
| [9] | *Carassius auratus* | Outdoor ponds | 4898 images | None, Weak, Medium, Strong | DL-MobileViT-SENet | 97.95% |
| [102] | -- | Laboratory | 720 images | None, Weak, Strong | Improved YOLOv8 | 83.33% |
| [99] | *Micropterus salmoides* | Laboratory | 900 videos | Low, Medium, High | YOLOv8-ByteTrack + ST-GCN | 98.47% |
| [109] | -- | Laboratory | 31635 images | None, Weak, Medium, Strong | GCVC | 93.33% |
| [96] | *Oncorhynchus mykiss* | Laboratory | 2685 images | Weak, Moderate, Strong | CFFI-Vit | 98.63% |
| [110] | *Cyprinus carpio* | Outdoor ponds | 452 videos | None, Weak, Medium, Strong | MIPS-EfficientNet | 97.00% |
| [104] | *Ctenopharyngodon idella, Carassius auratus* | Outdoor ponds | 889 videos | None, Weak, Strong | MobileViT-CBAM-BiLSTM | 98.61% |
| [111] | *Ctenopharyngodon idella* | Outdoor ponds | 18486 images | None, Weak, Strong | AKEM-RCNN | 96.94% |
| [11] | *Lateolabraxjaponicus* | Indoor factory farming | 5742 images | None, Weak, Medium, Strong | DCA-MVIT | 83.13% |
| [40] | *Tilapia* | Indoor factory farming | 7600 images | None, Weak, Medium, Strong | ResNet-MoVIT-ConvNeXt | 99.19% |
| [112] | *Salmoides* | Indoor factory farming | 850 videos | None, Weak, Strong | MFE-MobileViTv3 | 96.70% |
| [44] | *Hybrid groupers* | Laboratory | 2212 images | Satiation: 20%, 40%, 60%, 80%, 100% | TinyYOLO-poses + GCN | 98.10% |
| [97] | *Largemouth bass* | Indoor factory farming | 4365 images | Weak, Medium, Strong | MobileViT-CoordAtt | 97.18% |
| [10] | *Micropterus salmoides* | Indoor factory farming | 11662 images | Starved; Not starved | YOLOv8s-ALL | 95.21% |
| [41] | *Tilapia* | Indoor factory farming | 797 videos | None, Weak, Medium, Strong | STCA-MobileViTv3 | 96.88% |
| [46] | *Hybrid grouper* | Laboratory | 20000 images | 10-point scoring scales | CS-TransNeXt | 95.25% |
| [113] | *Oncorhynchus mykiss* | Laboratory | 299 videos | None, Medium, Strong | Fishsort + MLP | 96.61% |
| [114] | *Cyprinus rubrofuscus* | Laboratory | 4592 images | None, Weak, Strong | Optimized YOLOv8n | 94.70% |

Note: MKEM (Modified Kinetic Energy Model); RNN (Recurrent Neural Network); CA (Coordinate Attention); MSIF (Multi-Scale Information Fusion); DL (Dual Label); GCVC (Graph Convolution Vector Calibration); MIPS (Multi-step Image Pre-enhancement Strategy); AKEM-RCNN (Advanced Kinetic Energy Model Combined with the Recurrent Convolutional Neural Network); DCA-MVIT (DSGated convolution and Coordinate Attention for Multiscale Vision Transformers version 2); MFE (Multi-Feature Extraction); MTAFFB (Multi-Task Active learning framework); MLP (Multilayer Perceptron)

## 3.2 Fish feeding behavior analysis based on acoustic technology

Compared with computer vision methods, acoustic methods are insensitive to optical imaging conditions such as illumination changes, water turbidity, fish occlusion, and water surface reflection. Therefore, they are particularly suitable for aquaculture environments with low light, complex water conditions, or limited visual monitoring. Acoustic methods usually rely on devices such as hydrophones, acoustic tags, or sonar systems to capture feeding-related acoustic signals, echo patterns, and soundscape changes caused by the fish school activity during feeding [13,63,115]. These acoustic signals can reflect fish feeding status and feeding intensity from the perspectives of time, frequency, energy, and spectral structure, providing an important information source for non-contact behavior monitoring. Figure 4 shows the process of fish feeding behavior analysis based on acoustic techniques.

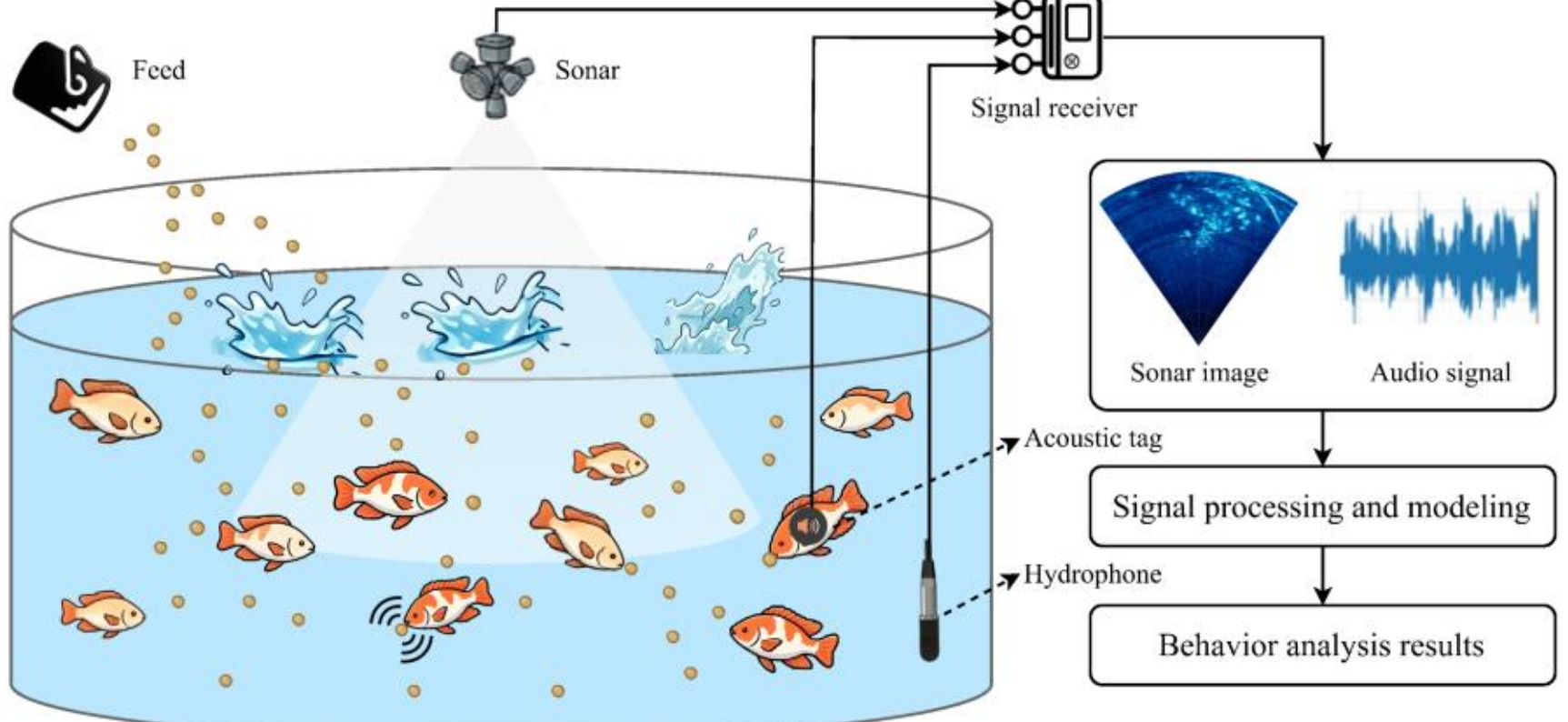

Figure 4. A framework for analyzing fish feeding behavior based on acoustic technology.

### 3.2.1 Fish feeding behavior recognition based on acoustic technology

Early studies mainly focused on the spectral characteristics of feeding-related acoustic signals and their relationship with behavioral responses. Feeding acoustic signals from different fish species typically concentrate within specific frequency bands, such as turbot (7–10 kHz) [116], rainbow trout (0.02–25 kHz) [117], and Japanese minnow (1–10 kHz) [118]. Mallekh et al. [119] further established a linear relationship between acoustic signals and feeding behavior, noting that the 6–8 kHz frequency range effectively reduces background noise interference. These studies indicate that acoustic signals in specific frequency bands can serve as important cues for feeding behavior recognition. However, such frequency band features are usually strongly dependent on species and scenarios, which limits their applicability across different environments.

On this basis, acoustic methods have gradually been extended to long term dynamic monitoring of feeding activity, such as feeding behavior recognition and movement tracking through acoustic telemetry systems [120–122]. Compared with short duration sound event recognition, acoustic telemetry is more suitable for recording long term changes in individual or group behavior. However, it usually requires fish to carry externally attached or implanted acoustic tags, and therefore faces challenges in cost, fish interference, and representativeness in scaled applications. In addition, passive acoustic methods recognize feeding behavior by analyzing chewing sounds, swallowing sounds, and soundscape metrics, such as the acoustic complexity index and acoustic entropy [63,123]. These methods broaden the application scope of acoustic monitoring and are especially suitable for underwater environments where clear images cannot be obtained stably. However, their recognition results are easily affected by sound sources unrelated to feeding, such as aerators, pumps, feeders, bubbles, and environmental background noise.

The above studies verify the feasibility of acoustic signals for feeding behavior recognition, but most of them remain at the level of correlation analysis and behavior monitoring, with limited automation and recognition accuracy. To improve performance, recent studies have gradually introduced artificial intelligence methods for data driven modeling of acoustic features. Liu et al. [31] used low dimensional time and frequency features together with dimensionality reduction methods to achieve high accuracy in fish feeding behavior recognition. Bies et al. [124] constructed a Deep Neural Network (DNN) recognition model based on splash sounds, and its performance exceeds that of a traditional CNN architecture. To address the small acoustic differences among different fish behaviors and the difficulty of feature learning, Xu et al. [125] combined Mel-Frequency Cepstral Coefficient (MFCC) with ResNet to fuse multiscale features, effectively distinguishing feeding behavior from swimming behavior. However, this method is limited by sample size and network redundancy. To address these limitations, Yang et al. [126] constructed a larger-scale dataset and introduced an improved SEResNet with Temporal Aggregation and Pooling (TAP) mechanisms, which further improves recognition accuracy while maintaining high efficiency. Table 3 summarizes representative studies on fish feeding behavior recognition based on acoustic techniques.

Table 3. Representative works on fish feeding behavior recognition based on computer vision.

| References | Fish species | Scenarios | Dataset size | Behaviors | Models | Evaluation metrics |
|---|---|---|---|---|---|---|
| [125] | *Sebastes schlegelii* | Laboratory | 812 | Swimming, Feeding | MFCC + ResNet152 | Accuracy = 99.00% |
| [124] | *Dicentrarchu labrax* | Aquaculture farm | 53 | Fish feeding splashing, Pellets rolling, Recirculating filter sound, Ambient noise | DNN | F1-score = 96.00% |
| [63] | *Salmo salar* | Indoor factory farming | -- | Pellets delivery, Feeding | Correlation analysis | -- |
| [31] | *Micropterus salmoides* | Laboratory | 753 | Swallowing, Chewing | RF | Accuracy = 98.63% |
| [126] | *Oncorhynchus mykiss* | Laboratory | 1020 | Swimming, Feeding, Jump | TAP-SEResNet | Accuracy = 98.17% |

### 3.2.2 Fish feeding intensity quantification based on acoustic technology

Fish in different hunger or satiety states usually show clear differences in chewing sounds, swallowing sounds, splash sounds, and overall sound field changes induced by feeding activity. Therefore, quantitative assessment of feeding activity and feeding demand can be achieved by analyzing the time domain and frequency domain features of acoustic signals. Cao et al. [127] used passive acoustics to collect feeding sound signals from individual largemouth bass and selected key feature parameters to measure feeding activity. This method can establish a relatively intuitive relationship between acoustic features and feeding intensity, but it usually relies on manual feature design and empirical judgment, and has limited adaptability to complex noisy scenes and different farming species. Cui et al. [128] converted audio signals into Mel Spectrograms (MS) and used a CNN to quantify fish feeding intensity. However, CNNs still have certain limitations in modeling global spectrogram structures and capturing long term temporal dependencies, which makes it difficult to fully represent the dynamic process of feeding intensity changes over time.

To enhance the ability of models to represent key frequency bands and complex spectrogram patterns, attention mechanisms and Transformer structures have gradually been introduced into feeding intensity quantification tasks. Huang et al. [129] introduced an attention mechanism and SimAM fusion module in the VGG16 framework to enhance the representation ability of spatial and channel features. Zeng et al. [130] proposed the Audio Spectrum Swin Transformer (ASST) model based on the attention mechanism, achieving an accuracy of 96.16% in the fish feeding intensity quantification task. However, this method has difficulty distinguishing changes in feeding intensity and spectrogram features in fry farming scenarios or fish species with very weak feeding sounds.

Lightweight networks and multi-feature fusion methods are used to balance accuracy and deployment efficiency. By constructing multiple time-frequency feature maps and combining them with lightweight networks such as MobileNetV3-SBSC and LC-GhostNet, high-accuracy feeding intensity quantification can be achieved [14,62]. However, multi-feature fusion also increases preprocessing and model computation complexity. Iqbal et al. [15] introduced an Involutional Neural Network (INN), which models relational structures and self-attention information in the feature space, achieving high recognition accuracy while maintaining low computational cost. Li et al. [131] combined MFCC features with a frequency-time attention mechanism, and introduced GhostConv and frequency-prioritized convolution to reduce computational complexity while strengthening the modeling ability for key frequency bands. This method shows good noise resistance and real time performance, and is suitable for deployment on embedded devices. Table 4 summarizes representative works on fish feeding intensity quantification based on acoustic technology.

### 3.2.3 Summary

Acoustic methods capture temporal and spectral signals during fish feeding, providing an important information source for fish feeding behavior analysis that differs from visual methods. The technological evolution is reflected in research paradigm, feature representation, and model structure. Early studies rely on specific frequency bands and empirical thresholds to detect feeding sounds. These methods have strong interpretability, but their degree of automation and cross scenario adaptability remain limited. With the introduction of artificial intelligence methods, acoustic approaches can automatically extract MFCC, MS, and multiscale time and frequency features, thereby capturing subtle differences in feeding behavior. Model structures also gradually evolve from conventional convolutional networks to deep architectures that integrate attention mechanisms and lightweight designs, balancing accuracy and deployment feasibility. Although acoustic methods are not affected by illumination and can adapt to complex water environments, aquaculture farms contain many non-feeding sound sources, including aerators, pumps, feeding devices, water flow, bubbles, rainfall, and background biological sounds. These noises may overlap with feeding sounds and reduce recognition reliability. Feeding sounds show clear dependence on species, body size, feed type, and farming density. Moreover, frequency ranges, intensity labels, and evaluation criteria are not yet unified across studies, which limits result comparison and model transfer. In some fish species or fry stages, feeding sounds are weak, making acoustic features difficult to extract stably. The placement of hydrophones, device sensitivity, and farming system structure also affect sound propagation and signal quality.

Table 4. Representative works on fish feeding intensity quantification based on computer vision.

| References | Fish species | Application scenarios | Dataset size | Intensity level | Models | Accuracy |
|---|---|---|---|---|---|---|
| [128] | *Oplegnathus punctatus* | Indoor factory farming | 3000 | None, Weak, Medium, Strong | CNN | 65.50% |
| [130] | *Oncorhynchus mykiss* | Laboratory | 24000 | None, Weak, Medium, Strong | ASST | 96.16% |
| [14] | *Oplegnathus punctatus* | Indoor factory farming | 9284 | None, Medium, Strong | MobileNetV3-SBSC | 85.90% |
| [132] | *Oplegnathus punctatus* | Indoor factory farming | 5447 | None, Weak, Medium, Strong | LC-GhostNet | 97.94% |
| [15] | *Oplegnathus punctatus* | Indoor factory farming | 9206 | None, Medium, Strong | INN | 97.11% |
| [129] | *Tilapia* | Laboratory | 4186 | None, Weak, Medium, Strong | VGG16-fish | 94.37% |
| [131] | *Micropterus salmoides* | Laboratory | 2221 | None, Medium, Strong | FTPG-CNN6 | 97.30% |

Note: LC (Learned Group Convolutions and Coordinated Attention).

## 3.3 Fish feeding behavior analysis based on sensors

Sensor methods monitor and evaluate feeding status and feeding demand by directly or indirectly sensing physiological, kinematic, and environmental changes during fish feeding. Unlike methods that rely on external images or acoustic signals, sensors can provide continuous and stable data sources, and show good environmental adaptability under low light conditions, turbid water, visual occlusion, or complex acoustic noise.

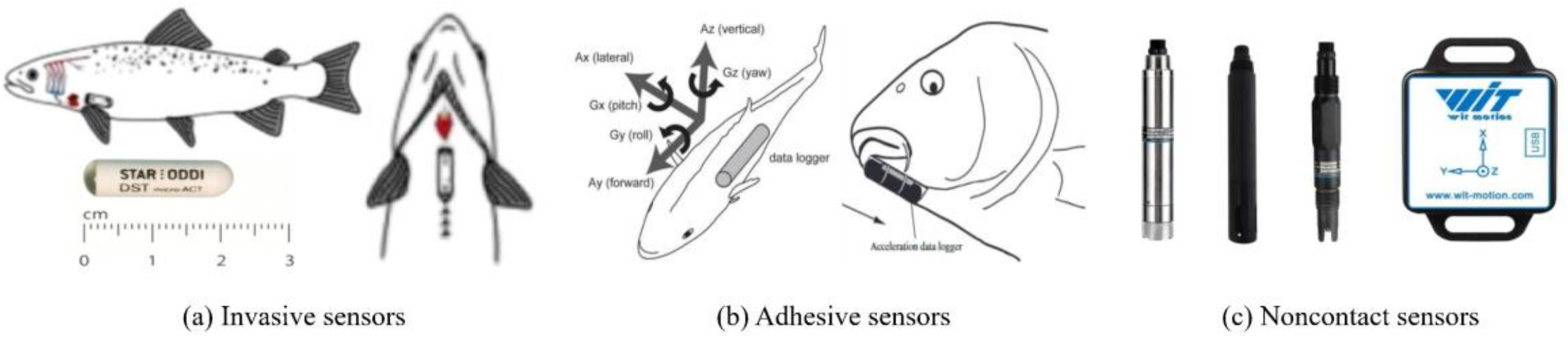

(a) Invasive sensors (b) Adhesive sensors (c) Noncontact sensors

Figure 5. Common sensors for fish feeding behavior analysis.

According to deployment mode and sensing object, sensors can be divided into three categories: invasive sensors, attached sensors, and external noncontact sensors [19,133,134]. Invasive sensors, such as implanted heart rate monitors and electromyographic sensors, can capture microscopic physiological responses of individual fish with high precision. Attached sensors, such as triaxial accelerometers fixed to the fish body or mandible, can record individual feeding actions, posture changes, and motion patterns. External noncontact sensors, such as water quality probes, suspended inertial sensors, flow velocity sensors, and vibration sensors, can monitor water surface fluctuations, flow disturbances, and changes in environmental parameters without directly interfering with fish bodies, and are therefore more suitable for long term group level aquaculture monitoring. Common sensor types are shown in Figure 5.

### 3.3.1 Fish feeding behavior recognition based on sensors

During feeding, fish usually show increased heart rate, changes in body temperature, enhanced mandibular movement, accelerated swimming, posture changes, and acceleration fluctuations [134]. These physiological and kinematic signals provide an important basis for feeding behavior recognition. Clark et al. [59] studied the heart rate and internal temperature changes of bluefin tuna during free swimming and feeding states using implanted biosensors, finding that these two indicators remained higher than baseline levels during feeding. Cubitt et al. [135] used electromyographic transmitters to automatically distinguish feeding states in rainbow trout. These studies directly reveal the relationship between feeding behavior and individual physiological responses, and therefore have strong biological interpretability.

Motion sensors have high sensitivity in capturing feeding related action patterns, and are particularly suitable for individual behavior recognition and behavioral mechanism studies. Makiguchi et al. [61] attached a miniature triaxial accelerometer to the mandible of carp, and remotely monitored mandibular movement, observing that the vibration frequency and amplitude during feeding were significantly higher than during non-feeding phases. Broell et al. [136] used high frequency accelerometer tags to study different behavioral patterns of fish, and showed that multiple parameters in the time domain and probability domain can statistically distinguish feeding behavior from escape behavior. Adegboye et al. [51] further used a data logger containing a triaxial accelerometer, a magnetometer, and a gyroscope to obtain acceleration and angular velocity data, and built a high accuracy recognition model for fish feeding behavior based on these data.

Although implanted or attached sensors can provide high quality individual behavior data, they may cause potential harm to fish, and individual based behavioral test results are difficult to use as true representations of group behavior. To improve the applicability of group monitoring, some studies have begun to explore external noncontact sensors. Subakti et al. [70] used a sensor suspended on the water surface to sense changes in water ripple acceleration caused by fish activity, observing three unique patterns related to fish feeding activities, which provided new insights for recognizing fish feeding behavior based on external sensors. Compared with individual sensors, external noncontact sensors cause less disturbance to fish and are more suitable for long term group monitoring. However, the collected signals are typically the result of the combined effects of fish school activity, water column responses, and environmental disturbances. Therefore, filtering, feature extraction, and multisource information fusion methods need to be further integrated to improve the reliability of feeding behavior recognition. Table 5 summarizes representative works on fish feeding behavior recognition based on sensors.

Table 5. Representative works on fish feeding behavior recognition based on sensors.

| References | Fish species | Application scenarios | Parameters | Behaviors | Models | Accuracy |
|---|---|---|---|---|---|---|
| [135] | *Oncorhynchus mykiss* | Laboratory | Electromyogram | Feeding, Non-feeding | SVM | 85.00% |
| [61] | *Cyprinus carpio* | Laboratory | Surging acceleration | Feeding, Non-feeding | -- | 92.82% |
| [136] | *Myoxocephalus polyacanthoceaphalus* | Laboratory | Acceleration | Feeding, Escape | DPA | 80.00% |
| [70] | *Oreochromis niloticus* | Outdoor ponds | Acceleration caused by water wave | Idle, Spike, Feeding | -- | -- |
| [51] | *Seriola quinqueradiata, Thynnus thynnus* | Laboratory | Acceleration, Angular velocity | Feeding, Escape, Routine movements, Swimming | MLPNN | 100% |
| [137] | -- | Laboratory | Electromyogram | Feeding, Non-feeding | Bayesian | 85.00% |

Note: DPA (Discrete Parameter Analysis); MLPNN (Multilayer Perceptron Neural Network).

### 3.3.2 Fish feeding intensity quantification based on sensors

Fish feeding activity is significantly coupled with water quality indicators such as dissolved oxygen, water temperature, and ammonia nitrogen [19]. For example, feeding activity can cause a local decrease in dissolved oxygen concentration, while changes in dissolved oxygen can in turn affect fish appetite and feed intake. Therefore, continuous changes in water quality parameters can reflect fish feeding intensity to some extent. Wu et al. [65] detected the appetite of silver perch based on dissolved oxygen changes during the feeding process and constructed an Adaptive Network-based Fuzzy Inference System (ANFIS) for feeding decision-making. Zhao et al. [66] further introduced water temperature as an additional input and optimized the parameters and fuzzy rule base using a hybrid learning method, showing significantly better prediction performance compared to traditional fuzzy logic control and manual feeding methods. Chen et al. [42] combined temperature, dissolved oxygen, and fish biomass to establish a mapping relationship between fish intake and environmental factors, avoiding the subjectivity of traditional empirical judgment. It should be noted that water quality parameters are usually indirect representation signals, and their changes are not completely determined by feeding activity. Aeration, water exchange, farming density, illumination, microbial metabolism, and environmental disturbance can all affect indicators such as dissolved oxygen, water temperature, and ammonia nitrogen. Therefore, judging feeding intensity based solely on water quality parameters is vulnerable to interference from non-feeding factors. In practical applications, water quality sensors are more suitable as auxiliary information for feeding demand assessment, environmental constraint judgment, and risk warning, rather than as the only basis for determining feeding intensity.

In addition to water quality parameters, motion and environmental sensors have also been used for feeding intensity quantification. Solpico et al. [17] developed a prototype sensor suite composed of water current sensors, cameras, and inertial measurement units to measure water flow around cages and record fish activity. The study found that when feeding started, the current increased continuously, and when feeding stopped, it dropped to zero. Solpico et al. [67] also conducted a study using water flow sensors for long-term monitoring of fish foraging activity. By measuring the water flow disturbance caused by fish movement, feeding intensity could be inferred without visual observation. These methods can reflect group feeding activity from the perspective of hydrodynamic responses, and are suitable for scenarios with limited visibility, difficult individual separation, or high-density farming.

With the development of multi-axis inertial sensors and deep temporal models, feeding intensity quantification has gradually shifted from single-signal analysis to multisource signal fusion. Ma et al. [18] introduced a six-axis inertial sensor to include angular velocity and angle data, and proposed a Time-domain and Frequency-domain Fusion model (TFFormer), which effectively reduces interference from device vibration, fish overlap, water turbidity, and complex illumination. However, this method still faces certain challenges in fry or low-density farming scenarios. Zhang et al. [138] combined triaxial vibration signals with deep learning to establish a dynamic prediction model for fish feeding intensity, which is significantly better than traditional optical flow methods and graph convolutional network based methods. Table 6 summarizes representative works on fish feeding intensity quantification based on sensors.

Table 6. Representative works on fish feeding intensity quantification based on sensors.

| References | Fish species | Application scenarios | Parameters | Intensity level | Models | Evaluation metrics |
| --- | --- | --- | --- | --- | --- | --- |
| [65] | *Bidyanus bidyanus* | Laboratory | Dissolved oxygen | Lack of appetite, Strong appetite | ANFIS | Accuracy = 97.89% |
| [66] | *Ctenopharyngodon idellus* | Outdoor ponds | Dissolved oxygen, Water temperature | Feeding percent | ANFIS | RMSE = 0.054 |
| [42] | *Oncorhynchus mykiss* | Laboratory | Water temperature, Dissolved oxygen, Weight, Number | Feed intake | MEA-BPNN | RMSE = 6.890 |
| [18] | *Lateolabrax japonicus* | Laboratory | Acceleration, angular velocity and angle caused by water wave | None, Weak, Medium, Strong | TFFormer | Accuracy = 91.52% |
| [138] | *Largemouth bass* | Indoor factory farming | Three-axis displacement signals | Feeding intensity | LSTM | RMSE = 69.43 μm |

Note: MEA (Mind Evolutionary Algorithm).

### 3.3.3 Summary

Sensor-based methods perceive feeding status and feeding intensity by directly or indirectly measuring fish physiological responses, motion patterns, and environmental parameters. Their technical evolution is mainly reflected in sensing objects, sensing modes, and task objectives. Early studies mostly rely on single signals such as heart rate, body temperature, electromyography, mandibular vibration, or acceleration for feeding behavior analysis. These methods have high accuracy and strong interpretability, but they are mainly designed for individual monitoring. In recent years, studies have gradually introduced multisource heterogeneous information, including angular velocity, posture angle, water quality parameters, biomass, water flow changes, and vibration signals, to enhance the stability of group feeding state recognition and intensity quantification. The sensing mode has also expanded from invasive individual monitoring to noninvasive group monitoring, improving the applicability of sensor-based methods in large scale aquaculture scenarios. The task objective has further extended from simple behavior judgment to the prediction of feeding demand and feed intake, providing a basis for precision feeding decisions. Although sensor-based methods have advantages in continuous monitoring and adaptability to complex environments, their practical application still faces multiple challenges. Invasive and attached sensors may induce stress or alter natural behavior, and data from a small number of individuals are difficult to use as a reliable representation of group status. External noncontact sensors usually collect indirect signals, which are easily affected by water flow, aeration, device vibration, farming density, and environmental changes. Water quality parameters are not exclusive representations of feeding behavior, and using them alone can easily lead to misjudgment. In scenarios involving fry, low density farming, or weak feeding activity, water flow, vibration, and ripple signals may be unclear, which limits model performance.

### 3.4 Fish feeding behavior analysis based on multimodal fusion

Multimodal fusion integrates information from different data sources, such as images, videos, sounds, sensors, and textual information, to obtain more comprehensive and accurate perception results, thereby supporting richer behavior analysis and decision making [27,139–141]. In fish feeding behavior analysis, different modalities usually provide complementary representation capabilities. The visual modality offers rich cues related to spatial structure and behavioral motion, but is easily affected by environmental factors such as water turbidity, light reflection, and occlusion. The acoustic modality captures temporal and spectral features during feeding and is sensitive to feeding events, but it is vulnerable to environmental noise and non-feeding sounds. The sensor modality provides structured measurements of motion, physiology, or environmental conditions, and has relatively high deployment feasibility, but its behavioral specificity is comparatively limited. Due to the distinctiveness and complementarity of different modalities in terms of information sources, noise resistance, and behavioral representation, multimodal fusion has gradually become an important direction in fish feeding behavior analysis.

From the perspective of fusion levels, multimodal fusion can be divided into early, intermediate, late, and hybrid fusion strategies [142], as shown in Figure 6. Early fusion usually aligns and combines different modalities at the model input stage or shallow feature stage, allowing the model to learn correlations between modalities at an early stage. It is suitable for scenarios where modalities are highly synchronized, have similar sampling frequencies, and show clear physical relationships. Intermediate fusion first extracts latent features through modality specific networks, and then performs fusion in the embedding space through concatenation, attention mechanisms, gating mechanisms, or learned weights. This strategy balances the preservation of modality specific characteristics and semantic complementarity, and is therefore commonly used in current fish feeding behavior analysis. Late fusion trains separate models for different modalities, and integrates their prediction results at the decision level through weighting, voting, or probabilistic ensemble. It has good modular independence and fault tolerance, and is suitable for practical applications with heterogeneous sampling rates, missing modalities, or high requirements for system maintainability. Hybrid fusion combines multiple strategies above, and enables multistage information interaction across low level features, intermediate semantics, and high-level decisions. This approach helps enhance robustness in complex aquaculture environments, but usually increases model complexity and deployment cost.

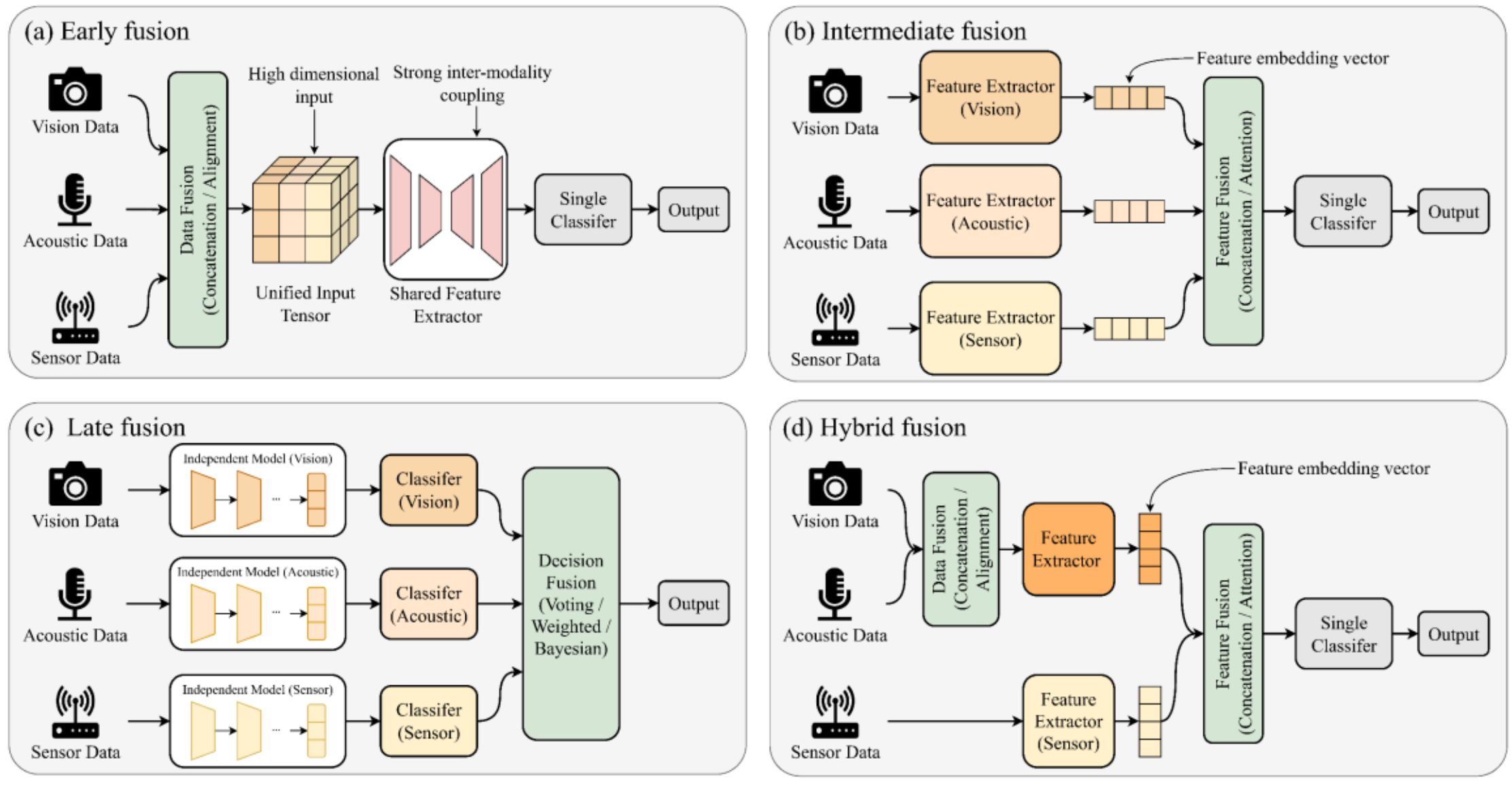

Figure 6. Multimodal fusion strategies categorized by integration stage.

### 3.4.1 Fish feeding behavior recognition based on multimodal fusion

Compared to single-modality methods, multimodal fusion can simultaneously exploit representation information of fish feeding behavior from visual, acoustic, and other sensing dimensions. Therefore, it shows higher robustness and environmental adaptability in behavior recognition tasks. Early studies focus more on validating correlations between different modalities than on joint recognition modeling in the strict sense. Qu et al. [143] used an audio-video synchronization method to study the relationship between feeding behavior and acoustic signals of largemouth bass under different farming densities. However, the video data is mainly used to extract the occurrence time of swallowing and swimming behaviors and to synchronize these events with audio signals, rather than being directly incorporated into the behavior recognition model. This study provides a basis for acoustic signal interpretation and cross-modal annotation, but remains limited in automated recognition and fusion modeling.

Table 7. Representative works on fish feeding behavior recognition based on multimodal fusion.

| References | Application scenarios | Modalities | Dataset size | Behaviors | Models | Accuracy |
|---|---|---|---|---|---|---|
| [35] | Laboratory | Image, Audio | 812 audios + 12284 images | Feeding, Swimming | U-FusionNet-ResNet50 + SENet | 93.71% |
| [144] | Laboratory | RGB video, Optical flow video | 3198 videos | Feeding, Normal, Hypoxia, Frightening, Hypothermia | DSC3D | 95.79% |
| [145] | Laboratory | Image, Audio | 1020 audios + 20033 images | Feeding, Swimming, Jump | Mul-SEResNet50 | 94.25% |

With the development of deep learning, multimodal fusion has gradually shifted from modality synchronization and correlation analysis to joint feature learning. Wang et al. [144] proposed the Dual-stream 3D Convolutional Neural Network (DSC3D) model to extract spatio-temporal features from RGB video and motion features from optical flow video, and fused them using a max-pooling strategy, achieving an average accuracy of 95.79% in fish feeding behavior recognition. This method demonstrates the advantage of fusing multiple visual cues in capturing appearance changes and motion information of fish schools, although the fusion mainly occurs within the visual modality. Further, Xu et al. [35] designed a multilevel fusion model based on acoustic and visual features to recognize swimming and feeding behaviors under complex conditions. This method fuses modality features at different stages through skip connection modules, enhancing information interaction between acoustic and visual modalities. To address the problem that blurred images and acoustic noise weaken modality complementarity in aquaculture environments, Yang et al. [145] introduced a multimodal interactive fusion module and a U-shaped bilinear fusion structure into ResNet50 to obtain a more comprehensive joint feature representation, thereby improving recognition accuracy without an obvious loss of detection speed. Table 7 summarizes representative works on fish feeding behavior recognition based on multimodal fusion.

### 3.4.2 Fish feeding intensity quantification based on multimodal fusion

Feeding intensity quantification requires not only determining whether fish are feeding, but also evaluating feeding activity level, sustained intensity, and dynamic change trends. This places higher requirements on the completeness of feature representation and fine-grained discrimination capability. Jointly using multisource information, such as video, audio, and sensor data, can more comprehensively represent group motion, acoustic changes, and environmental responses during fish feeding, thereby improving the accuracy and robustness of feeding intensity quantification. Syafalni et al. [146] processed video and accelerometer data using R(2+1)D convolutional layers and dense networks, respectively, achieving high accuracy on the validation set. Zheng et al. [147] used near infrared images and depth maps to jointly represent feeding intensity, and integrated feeding dynamics, water flow field changes, and related acoustic features through weighted fusion. Du et al. [20] developed a Multimodal Fusion of Fish Feeding Intensity (MFFFI) framework, which integrates deep features from audio, video, and sensor data, and outperforms unimodal methods. These studies show that multisource information fusion can improve feeding intensity quantification performance, especially when information from a single modality is incomplete or affected by environmental interference. However, simple weighted fusion or channel concatenation is often insufficient to fully exploit nonlinear complementarity among different modalities, and it is also difficult to dynamically adjust modality contributions according to environmental changes.

To address insufficient exploitation of modality complementarity, recent studies have introduced attention mechanisms, co-attention structures, and cross modal Transformers to enhance information interaction among different modalities. Yang et al. [148] combined feature compression, expansion, and separation strategies to enhance video and audio features, and designed a deep modular co-attention mechanism for intermodal interaction. Hu et al. [149] added a multimodal transfer module and an adaptive weighting mechanism to the MulT algorithm, enabling effective fusion of feature vectors and dynamic adjustment of modality contributions. Li et al. [22] proposed a two-stage attention fusion framework that captures the spatiotemporal complementarity between acoustic and visual modalities through a dimension alignment strategy and an attention mechanism. Wang et al. [150] designed a temporal connection module and a cross modal fusion module to achieve intersegment temporal correlation modeling and deep fusion of optical flow and RGB features, and introduced a selective state space model for efficient modeling of spatiotemporal consistency.

As multimodal models continue to increase in scale and computational complexity, balancing accuracy, efficiency, and deployment cost has become a new research focus. Gu et al. [21] developed an audio and video aggregation module composed of self-attention and cross-attention mechanisms, and introduced a lightweight separable convolutional feedforward module to achieve a balance between speed and accuracy. Zhang et al. [23] proposed a recognition framework based on multimodal fusion and enhanced knowledge distillation to reduce computational complexity and deployment cost. Cui et al. [151] introduced a unified hybrid modality method, U-FFIA, which uses models pretrained on single-modal data for knowledge distillation, thereby reducing the computational overhead caused by independent multimodal encoders. Table 8 summarizes representative works on fish feeding intensity quantification based on multimodal fusion.

### 3.4.3 Summary

Multimodal fusion methods integrate visual, acoustic, sensor, and other management data, providing a more comprehensive behavioral representation for fish feeding behavior analysis. Their technical evolution is mainly reflected in research paradigms, fusion levels, and research objectives. In terms of research paradigms, early studies mainly focus on modality synchronization, correlation validation, and simple feature combination, while recent studies gradually shift toward cross modal joint modeling and dynamic feature interaction. Fusion levels have evolved from simple concatenation to explicit interaction, and techniques such as attention mechanisms, co-attention structures, and cross modal Transformers effectively enhance modality information transfer and alignment. Research objectives have also expanded from optimizing accuracy alone to achieving a comprehensive balance among robustness, efficiency, and deployment adaptability. Although multimodal fusion usually outperforms unimodal methods, different modalities often differ in sampling frequency, data dimension, and time delay, making accurate synchronization and alignment difficult. Multimodal data acquisition systems usually involve multiple types of devices, and their deployment cost, maintenance cost, and system complexity are higher than those of unimodal solutions. In practical aquaculture processes, some modalities may be missing because of water turbidity, noise interference, sensor failure, or communication abnormalities, so models need robust inference capability under missing modality conditions. Redundant and noisy information may also exist across different modalities. Without an effective modality selection mechanism, fusion may instead introduce interference. Existing studies are mostly validated on specific fish species, specific farming systems, and relatively limited datasets. Their ability to generalize across fish species and farming environments, as well as their capacity for long term online operation, still requires further evaluation.

Table 8. Representative works on fish feeding intensity quantification based on multimodal fusion.

| References | Application scenarios | Modalities | Dataset size | Intensity level | Models | Accuracy |
|---|---|---|---|---|---|---|
| [152] | Laboratory | Image, Audio | 1770 | None, Weak, Strong | FC n-fused CNN | 96.50% |
| [149] | Laboratory | Water quality, Audio, Image | 1293 | None, Weak, Medium, Strong | Fish-MulT | 95.36% |
| [148] | Laboratory | Water quality, Video, Audio | 102 | None, Weak, Medium, Strong | DMCA-UMT | 75.30% |
| [153] | Laboratory | Video, Audio | 15700 | None, Weak, Medium, Strong | FCN-LSTM-ResNet | 99.81% |
| [146] | Laboratory | Video, Accelerometer | 2475 | Low, High | R(2+1)D-CNN | 99.09% |
| [20] | Indoor factory farming | Audio, Video, Image | 7611 | None, Weak, Medium, Strong | MFFFI | 99.26% |
| [147] | Laboratory | Audio, Depth map, Near infrared image | 13 | None, Medium, Strong | MFWF | 97.00% |
| [21] | Indoor factory farming | Video, Audio, Dissolved oxygen | 7800 | None, Weak, Medium, Strong | MMFINet | 97.60% |
| [154] | Indoor factory farming | Dissolved oxygen, Image | 10568 | None, Weak, Medium, Strong | MMKDR | 96.65% |
| [155] | Indoor factory farming | Video, Audio | 2330 | None, Weak, Strong | HGCM | 97.61% |
| [151] | Indoor factory farming | Video, Audio | 27000 | None, Weak, Medium, Strong | U-FFIA | 90.70% |
| [156] | Indoor factory farming | Video, Audio | 27000 | None, Weak, Medium, Strong | ACAF-Net | 95.40% |
| [157] | Indoor factory farming | Image, Audio, Water wave | 7089 | None, Weak, Strong | PMIN | 96.76% |
| [158] | Indoor factory farming | Sonar images, Audio | 7611 | None, Weak, Medium, Strong | LightHybridNet-Transformer-FFIA | 95.42% |
| [150] | Indoor factory farming | Video, Optical flow | 1920 | None, Weak, Medium, Strong | DSSFM | 95.52% |
| [159] | Laboratory | Video, Sensor | 7250 | None, Weak, Medium, Strong | VTFish | 94.29% |

Note: FCN (Fully Convolutional Network); MFWF (Multi-Feature Weighted Fusion); MMFINet (Multimodal Fusion Interaction Network); MMKDR (Multimodal Knowledge Distillation Recognition); HGCM (Cross-Modal Hierarchical Gating Network); FFIA (Fish Feeding Intensity Assessment); ACAF-Net (Adaptive Cross-modal Attention Fusion Network); PMIN (Progressive Multimodal Interaction Network); DSSFM (Dual-Stream Spatiotemporal Fusion Method).

# 4. Application status of fish feeding behavior analysis

## 4.1 Applications in intelligent feeding

Intelligent feeding is one of the most direct and central application scenarios of fish feeding behavior analysis. Traditional feeding mainly relies on manual experience-based judgment or fixed time, fixed ration mechanical feeding, and therefore cannot be dynamically adjusted according to the real time feeding status of fish schools and environmental changes. Such feeding modes easily lead to underfeeding or overfeeding, which in turn affects fish growth, feed utilization efficiency, water quality stability, and farming profitability. With the development of intelligent sensing, behavior analysis, and automatic control technologies, fish feeding is gradually shifting from experience driven and rule driven paradigms to data driven and closed loop control paradigms, as shown in Figure 7.

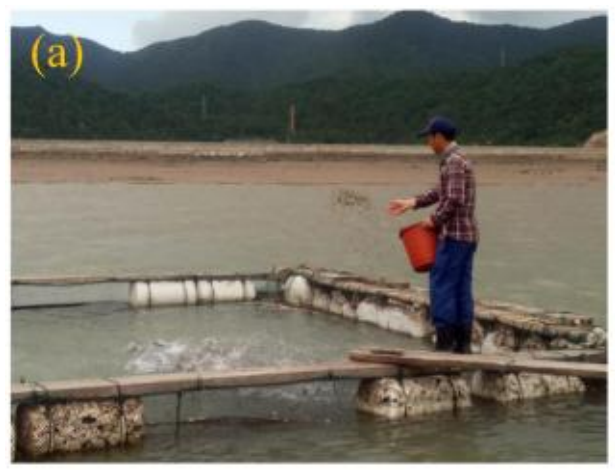

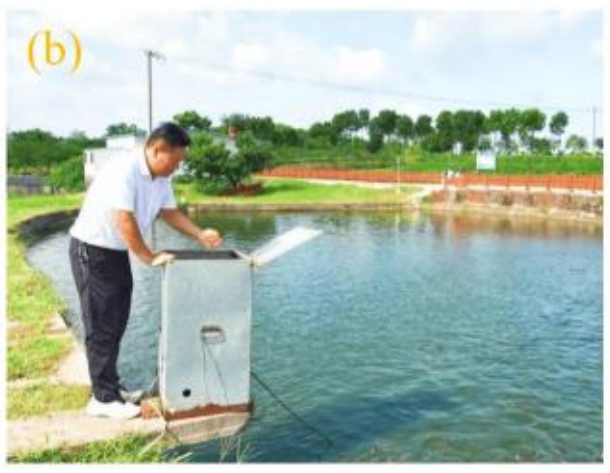

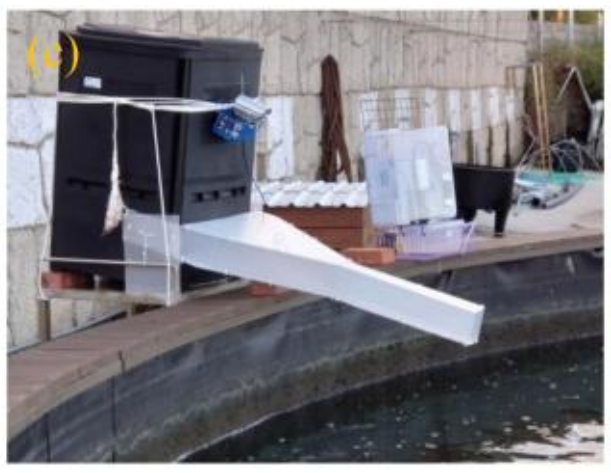

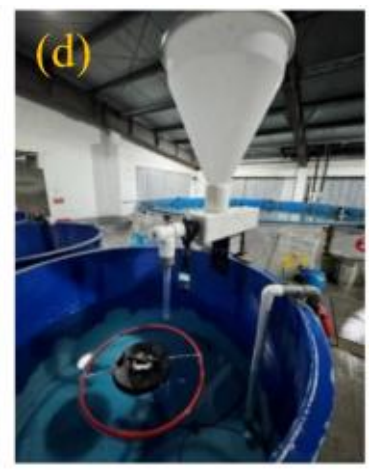

Figure 7. Feeding methods in aquaculture. (a) Manual feeding; (b) Mechanical feeding; (c) Computer vision-based intelligent feeding system [92] ; (d) Water wave sensor-based intelligent feeding system [138].

In intelligent feeding systems, fish feeding behavior analysis can directly support feeding decisions. Fish schools in a highly active feeding state usually exhibit rapid aggregation, intense competition for feed, frequent swallowing, stronger surface disturbance, or a clear increase in acoustic signals, whereas satiated fish schools are often characterized by reduced group activity, weaker aggregation, smoother swimming, and attenuated signals. By analyzing these behavioral and signal changes in real time, the system can determine whether the fish school still has feeding demand and then decide whether feeding should continue or stop. For example, Zhou et al. [160] extracted quantitative indicators of fish feeding behavior using Delaunay triangulation and image texture analysis, and then performed feeding decision-making through ANFIS, achieving an accuracy of 98%. Wang et al.[161] constructed a multi-task network to analyze fish foraging activities and monitor the number of uneaten feed particles, enabling the adjustment of feeding intervals and decision-making for feeding endpoints. Zhou et al. [162] introduced a deep learning-based object detection model to capture residual bait and feeding frequency, and then used a fuzzy neural network to derive the control strategy. To address the susceptibility of visual methods to water turbidity and imaging conditions, Hu et al. [92] determined whether feeding should continue by identifying water wave changes caused by fish feeding, showing good feeding decision capability in outdoor environments. Pan et al. [163] collected surface wave data using a six-axis sensor and selected representative feature parameters through PCA as the basis for feeding judgment, achieving an FCR of over 85%. To overcome the limitations of fixed-position feeding, Lu and Kong [164] designed a mobile autonomous feeding device that integrates a mechanical locomotion module, a feed distribution module, and a fish hunger monitoring module, enabling it to move across the culture area and perform distributed feeding at multiple points. This device achieves a fish recognition success rate of 92.3% and a feeding error below 10.67%, showing low feed loss and high operational reliability.

Fish feeding behavior analysis can also provide an important basis for dynamically adjusting the feeding amount in automatic feeding systems. Unlike traditional automatic feeding systems that rely only on preset programs, intelligent feeding systems based on behavior analysis can incorporate fish feeding status, residual feed amount, biomass, and environmental parameters into the decision process, thereby forming a closed-loop mechanism of perception, analysis, decision, and execution. The system developed by Aisuwarya & Suhendra [71] determines fish hunger status by monitoring water surface fluctuations, and dynamically adjusts the feeding amount according to fish body weight, so that the daily feed supply meets the ideal feeding standard. Zhang et al. [77] proposed a method for determining feed allocation by combining biomass estimation with fish feeding behavior recognition. Their system first evaluates the biomass in the culture area, then determines the initial feeding amount according to the feeding rate, and finally adjusts it through real-time monitoring. Yang et al. [165] used computer vision to obtain feeding frequency and residual bait information, and then applied a heuristic algorithm to adjust subsequent feed allocation, saving about 15% feed compared with manual feeding.

Although these methods promote the development of intelligent feeding, most systems still rely on preset rules, empirical thresholds, or relatively simple mathematical models, and therefore have difficulty fully adapting to complex, dynamic, and multifactor coupled aquaculture environments. In recent years, large language models, with their natural language understanding and reasoning capabilities, can simulate expert decision making and provide a new direction for feeding systems. Huang et al. [166] combined visual perception with a large language model to enable deep reasoning and adaptive optimization of feeding decisions, improving FCR by 73.24% and 32.60% compared with manual and fuzzy control methods, respectively. Karimanzira [167] integrated multi-sensor data with a multimodal vision-language model to achieve real-time monitoring and analysis of fish feeding behavior, while also adopting an adaptive strategy based on biomass, water quality, and fish behavior to adjust feed allocation, leading to a 53% improvement in FCR.

## 4.2 Applications in aquaculture management

Fish feeding behavior analysis can not only be used to optimize feeding strategies, but also provide an important behavioral basis for aquaculture management. Feeding behavior reflects the integrated response of fish to feed supply, environmental conditions, health status, and group vitality, and its changes often occur earlier than declines in growth performance or visible disease symptoms. Therefore, through long term, continuous, and fine-grained monitoring of feeding behavior, farmers can understand fish health status, environmental suitability, and system operation risks at the group level, thereby enabling more proactive aquaculture management.

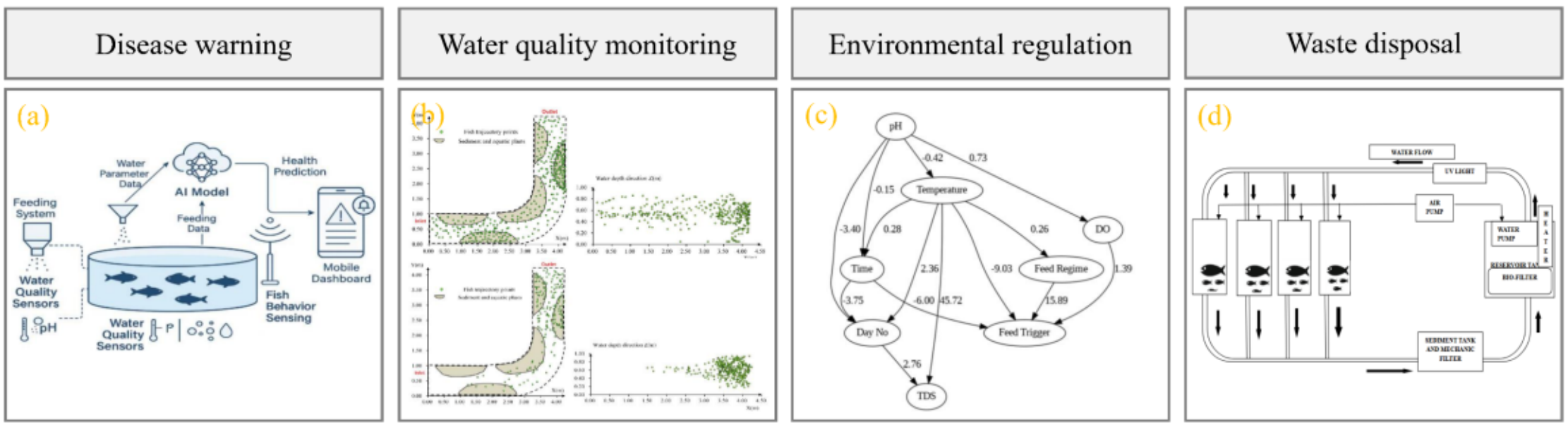

Figure 8. The application of fish feeding behavior analysis in aquaculture management.

Abnormal fish feeding behavior can serve as an important signal for health risk assessment and disease early warning. In the early stages of disease, stress, or environmental deterioration, fish usually show reduced appetite, disrupted feeding rhythms, weakened group activity, delayed feeding responses, or poor coordination between feeding and swimming behaviors. These changes often occur before obvious clinical symptoms appear. Therefore, continuous monitoring of feeding behavior helps detect potential health problems in advance, and provides a basis for farmers to adjust feeding, improve water quality, isolate diseased fish, or conduct further diagnosis. Egash & Teyene [168] developed a fish health prediction system by integrating water quality data, foraging behavior, and fish movement behavior, as shown in Figure 8(a). This system achieves an accuracy of 92.7% in abnormal state detection, provides early warnings before symptoms appear, and enables real time risk monitoring and management guidance through a mobile dashboard. Kedhareswari et al. [169] proposed an artificial intelligence driven Internet of Things system that identifies potential disease signs by analyzing behavioral patterns such as reduced activity, abnormal swimming, and changes in foraging habits, reducing fish mortality by approximately 10–20%. These studies show that combining feeding behavior with movement behavior, water quality parameters, and intelligent warning models can help shift fish health management from passive diagnosis to early prediction.

Fish feeding behavior analysis can also be used to evaluate the suitability of the farming environment. Fish are highly sensitive to environmental factors such as water temperature, dissolved oxygen, pH, ammonia nitrogen, turbidity, and pollutants. When environmental conditions deviate from suitable ranges, fish often show reduced feed intake, delayed feeding responses, abnormal activity levels, or changes in spatial distribution. Therefore, feeding behavior and related motion features can serve as indirect indicators of water quality changes and environmental stress. Existing studies show a close relationship between water quality deterioration and changes in feeding behavior, such as reduced feed intake, delayed feeding responses, and decreased group activity [19]. Yuan et al. [170] used computer vision to process and analyze video sequences of fish behavior in real time and extract motion parameters for water quality classification and early warning. Huang et al. [171] used acoustic tags to obtain fish movement trajectories and their spatiotemporal distribution during water quality changes, as shown in Figure 8(b), and established an aquatic ecosystem health assessment system whose results are highly consistent with conventional water quality indices. In addition to kinematic features, innate behavioral responses of fish are also used for refined early warning. For example, Gobiocypris rarus exhibits strong scototaxis. When exposed to cadmium ions ($Cd^{2+} \geq 3$ mg/L) or the anesthetic MS-222 ($\geq$11 mg/L), it exhibits disrupted light-avoidance behavior, which can serve as an early indicator of chemical pollution [172]. Low-turbidity water conditions can also induce significant physiological stress in some fish species, such as pikeperch, as reflected by slower feeding responses and elevated cortisol levels [173].

In intelligent environmental regulation, fish feeding behavior can provide a dynamic basis for decisions on aeration, water circulation, water exchange, and water quality adjustment [174]. Feeding activity is usually accompanied by enhanced metabolism, increased oxygen consumption, and a higher excretion load, while dissolved oxygen, water temperature, and water quality conditions in turn affect fish feeding motivation and feeding intensity. Therefore, a significant bidirectional coupling exists between feeding behavior and environmental parameters. Clear mathematical relationships have been established among oxygen demand, feeding rate, and water temperature. By analyzing feeding intensity in real time, it is possible to predict changes in dissolved oxygen over a future period and adjust aeration or water circulation in advance [175]. As aquaculture systems continue to expand in scale, environmental regulation faces the challenge of multi-objective optimization. Elmessery et al. [176] introduced a deep deterministic policy gradient algorithm to optimize control strategies and coordinate the relationship among feeding rate, dissolved oxygen, pH, and ammonia nitrogen. Palconit [177] used an AIoT system to collect long-term time-series data and employed a causal analysis tool to examine the effects of environmental factors on feeding behavior. The results showed that temperature makes a strong forward causal contribution to feeding behavior, whereas pH may affect feeding intensity indirectly through dissolved oxygen, as shown in Figure 8(c). Wang et al. [178] developed a multi-objective aquaculture environment control system that integrates a behavior recognition model, a specific growth rate model, and an energy cost model to balance economic benefit and fish welfare. In a 25-day experiment, the periodic profit achieved by this system was 1.03 to 8.93 times that of the conventional threshold control method, while ensuring no significant difference in growth relative to the control group.

A potential application of fish feeding behavior is to predict waste loads and optimize the operation of water treatment units in recirculating aquaculture systems. After feeding, uneaten feed and feces increase the loads of suspended solids, ammonia nitrogen, and nitrite in the system [179], thereby imposing higher demands on water treatment units such as biofilters and circulation pumps. When the system identifies that the fish school has entered a high-intensity feeding stage, a variable-flow regulation model can be used to increase the operating frequency of the circulation pump and accelerate water transport to the biofilter, so that ammonia nitrogen and nitrite can be removed before substantial accumulation occurs [180], as shown in Figure 8(d). In addition, studies based on system dynamics models show that coordinating feeding frequency with the nitrification rate of the biofilter reduces nitrogen accumulation by more than 35% and significantly improves the long-term stability of recirculating aquaculture systems [181]. An emerging application is to use fish feeding behavior to indirectly manage pollutants in the water body. Existing studies show that fish feeding activity may alter the surface properties of microplastics, enhance bacterial colonization, and affect the degradation of microplastics in aquatic environments [182]. This finding suggests that fish feeding behavior may participate in the migration and transformation of pollutants in aquaculture water. However, this direction is still at an exploratory stage, and its underlying mechanisms, applicable conditions, and management value still require further validation.

# 5 Challenges and future directions

Although fish feeding behavior analysis has made significant progress in task definition, technical support, and application exploration, the field remains in a transitional stage from experimental research to production application. Real aquaculture environments are highly dynamic and uncertain, and substantial differences exist among fish species, farming modes, equipment conditions, and management objectives. These factors still impose many limitations on the stability, generalization ability, and application value of existing methods in complex scenarios. From a perspective that integrates technical development and practical application, this paper summarizes the main challenges currently faced by fish feeding behavior analysis and outlines future research directions.

### 5.1 Robust perception in complex aquaculture environments

Most existing studies achieve good performance in laboratory or semi controlled environments, but interference factors in real aquaculture scenarios are more complex and clearly dynamic, which can easily lead to model performance degradation or even failure. Water turbidity, water surface reflection, bubble interference, fish occlusion, and illumination changes all affect image quality and the stable extraction of behavioral features. Especially under high density farming conditions, blurred individual boundaries and fish overlap further increase the difficulty of analysis. Acoustic methods do not depend on illumination conditions or water transparency, but they are highly sensitive to noise and therefore impose strict requirements on deployment environments. Signal separability is especially low in low density farming or when feeding sounds are weak. Sensor based methods can alleviate some limitations of visual and acoustic methods, but their outputs are still affected by fish density, individual differences, and deployment modes, leading to insufficient stability and interpretability. Therefore, future research needs to further enhance robust perception in complex environments. On the one hand, feature enhancement and noise suppression methods should be developed for weak signals, strong noise, and severe occlusion, so as to improve the stability of visual, acoustic, and sensor data in real scenarios. On the other hand, datasets that more closely reflect practical production conditions should be constructed to avoid models performing well only under short term, single scenario, or ideal data conditions. In addition, the evaluation of robust perception should not be limited to algorithmic accuracy, but should also comprehensively consider false alarm rate, missed detection rate, response latency, and system stability during long term online operation, thereby providing more reliable support for the engineering application of fish feeding behavior analysis.

### 5.2 Cross-species and cross-scenario generalization

Most existing studies focus on specific fish species and single farming environments, and their model designs are often strongly scenario dependent, making them difficult to transfer to other fish species or farming modes. Different fish species differ markedly in feeding patterns, swimming habits, aggregation modes, feeding sound characteristics, and environmental sensitivity. For example, surface-feeding and bottom-feeding fish show different patterns in water surface disturbance, spatial distribution, and feed response. Fish fry, juveniles, and adults also differ clearly in feeding intensity, movement amplitude, and acoustic signal strength. Meanwhile, differences among farming modes, such as ponds, cages, flow-through systems, and recirculating aquaculture systems, in water conditions, spatial structure, water exchange mode, and farming density further increase the uncertainty of feeding behavior representation and the difficulty of cross scenario transfer. In addition, existing datasets have limited scenario coverage and lack unified acquisition protocols and annotation standards. This problem is particularly evident in feeding intensity quantification, where intensity levels, scoring criteria, and label definitions vary across studies. Such inconsistency makes fair comparison of model performance difficult and limits the transfer of related methods to new scenarios. Therefore, future research can be advanced from both data and method perspectives. At the data level, standardized datasets covering multiple fish species, growth stages, and farming modes should be constructed, and unified protocols for data acquisition, label definition, and evaluation should be gradually established. At the method level, techniques such as transfer learning, domain adaptation, few-shot learning, unsupervised learning, and continual learning can be introduced to reduce model dependence on large scale manual annotation and improve adaptability to new fish species and new farming environments.

### 5.3 Multimodal collaborative modeling and lightweight deployment

Multimodal fusion can improve the robustness and granularity of fish feeding behavior analysis, but it also substantially increases the complexity of data processing, model development, and engineering deployment. Different modalities differ markedly in sampling frequency, temporal scale, data dimension, noise type, and physical meaning. Therefore, temporal synchronization, feature alignment, and information complementarity are core challenges in multimodal fusion. In real aquaculture environments, missing modalities, signal degradation, and noise accumulation further increase the uncertainty of fusion modeling. In addition, multimodal deep models usually have large parameter scales and high computational costs, which impose higher requirements on hardware devices and operating environments. Practical aquaculture scenarios require not only high recognition accuracy, but also real time response, low power operation, low maintenance cost, and long-term stability. If a model relies heavily on high performance computing devices or has large inference latency at the edge, its engineering application value will be limited. Therefore, future research needs to balance fusion performance and deployment efficiency. On the one hand, robust fusion mechanisms that can adapt to modality heterogeneity, missing modalities, and noise disturbance should be designed, such as dynamic weight allocation, modality selection, cross modal attention, and missing modality compensation. On the other hand, model pruning, knowledge distillation, lightweight networks, and edge computing should be combined to reduce computational cost and improve deployment feasibility in intelligent feeding devices, aquaculture monitoring terminals, and Internet of Things systems.

### 5.4 From behavior analysis to intelligent feeding

How to transform behavior analysis results into stable, reliable, and explainable feeding control strategies is a key issue for moving this field from algorithmic research toward production application. At present, many studies still remain at the stage of feeding state recognition, behavior classification, or intensity level judgment, and have not yet established clear and stable mappings with feeding control strategies. The relationship between feeding behavior and actual feeding demand is not a simple correspondence, but is jointly affected by multiple factors, including fish species, growth stage, hunger level, farming density, feed type, and feeding history. Under different environmental conditions, similar feeding intensity may correspond to different feeding demands, and the same feeding amount may also lead to different growth outcomes and feed utilization efficiency. Therefore, feeding decisions based only on a single behavioral indicator can easily result in underfeeding or overfeeding. An ideal intelligent feeding system should form a dynamic closed loop that integrates perception, analysis, decision making, execution, and feedback. Such a system should incorporate feeding behavior, environmental status, biomass, growth performance, and historical feeding records into a unified decision framework, enabling real time adjustment and continuous optimization of feeding strategies. Future research should promote the transition of fish feeding behavior analysis from perception driven modeling to decision driven optimization, and further reveal the quantitative relationships among behavioral features, feeding demand, feed utilization efficiency, and growth response. Meanwhile, feeding behavior analysis should be deeply coupled with growth models, environmental regulation models, feed consumption models, and automatic feeding devices, so as to construct intelligent closed loop feeding systems that can operate online over long periods, update parameters adaptively, and adapt to different farming scenarios.

### 5.5 Multitask collaboration and scalability

Fish feeding behavior not only reflects feeding demand, but is also closely related to fish health, water quality changes, metabolic waste accumulation, and system energy consumption. Therefore, feeding behavior analysis should not be used only as an auxiliary tool for intelligent feeding, but should be further developed into a key perception and decision node in precision aquaculture management. At present, tasks such as feeding control, health monitoring, and environmental control mostly operate in relatively independent ways, with limited cross task information sharing and collaborative decision making. This makes it difficult to achieve overall optimization at the system level. In practical aquaculture, different management objectives often involve both synergy and balance. For example, increasing the feeding amount may promote short term growth, but it can also increase residual feed accumulation, ammonia nitrogen load, and disease risk. Enhanced aeration and water circulation help improve water quality, but may also increase energy consumption and operating costs. Therefore, future research should promote the transition of aquaculture from single task management to multitask collaboration, so as to achieve a dynamic balance among feeding efficiency, fish growth, environmental stability, energy cost, fish welfare, and economic benefit. As aquaculture scale expands and intelligent devices increase, multisource data access, online model updating, collaborative equipment control, and remote management also impose higher requirements on system architecture. In the future, technologies such as the Internet of Things, edge computing, cloud platforms, and digital twins can be integrated to build scalable intelligent aquaculture systems, enabling collaborative operation of data acquisition, behavior analysis, environmental regulation, feeding control, and production evaluation. Meanwhile, system design should also consider long term stability, maintenance cost, fault diagnosis, and user operability, so as to prevent intelligent systems from remaining only at the stage of experimental validation.

### 5.6 Standardized Evaluation, interpretability, and long-term production validation

In addition to challenges at the algorithm and system levels, fish feeding behavior analysis still requires further improvement in standardized evaluation, model interpretability, and long-term production validation. At present, different studies vary substantially in data acquisition conditions, task definitions, data annotation, evaluation metrics, and experimental settings, which leads to limited comparability among research results. This problem is particularly evident in feeding intensity quantification and intelligent feeding applications, where intensity levels, scoring scales, definitions of feeding termination, and production outcome indicators have not yet been unified. Such inconsistency limits the comparative assessment and practical promotion of related methods. Therefore, it is necessary to establish a standardized evaluation system for fish feeding behavior analysis in future research, in which algorithmic performance, system performance, and production benefits are incorporated into a unified assessment framework.

In terms of model interpretability, deep learning and multimodal models have improved the performance of behavior recognition and intensity quantification, but their decision processes often lack transparency. For intelligent feeding and aquaculture management, the system should not only output results such as whether to feed, whether to reduce the ration, or whether to issue a warning, but also provide the corresponding decision basis, such as decreased group aggregation, weakened feeding sound energy, abnormal fluctuations in dissolved oxygen, or increased residual feed. Improving model interpretability not only helps enhance farmers' trust in intelligent systems, but also helps identify the causes of model errors in complex or abnormal scenarios.

In addition, most existing studies focus on short term experiments or small-scale validation, whereas real aquaculture production is characterized by long production cycles, large environmental variations, and complex management factors. Future research should strengthen long term validation across seasons, production batches, and commercial scale scenarios, and systematically evaluate the comprehensive effects of feeding behavior analysis techniques on feed utilization efficiency, residual feed control, water quality stability, growth performance, fish welfare, labor cost, and economic return. Only by combining model performance validation with real production outcome evaluation can the practical application value of fish feeding behavior analysis techniques in sustainable aquaculture be fully demonstrated.

## 6 Conclusion

This paper systematically reviews the task definition, technical support, application status, and future challenges of fish feeding behavior analysis in aquaculture. Overall, fish feeding behavior analysis has evolved from coarse-grained judgment based on manual experience and single features into a fine-grained analytical system that integrates multisource sensing information and focuses on dynamic processes. At the task level, related studies have expanded from simple feeding state recognition to feeding stage discrimination, feeding intensity quantification, feeding demand inference, and dynamic monitoring of the whole feeding process. At the technical level, computer vision, acoustic techniques, and sensor technologies show a transition from hand crafted features to data driven modeling, from static analysis to spatiotemporal modeling, and from unimodal analysis to multimodal fusion. At the application level, fish feeding behavior analysis has shown application potential in intelligent feeding, health warning, environmental regulation, and recirculating aquaculture system management, and is developing from a single behavior recognition tool into an integrated decision support system for aquaculture management.

Despite this progress, existing studies still face challenges such as insufficient robustness in complex aquaculture environments, limited cross scenario generalization, and high complexity and deployment difficulty in multimodal modeling. Meanwhile, the dynamic mapping between feeding behavior analysis results and feeding control, environmental regulation, and health management remains insufficient, and long-term online validation and production outcome evaluation are still relatively limited. In the future, fish feeding behavior analysis should move beyond improving single algorithmic performance and further focus on reliable perception, explainable decision making, and system level collaborative optimization for real aquaculture scenarios. On the one hand, standardized datasets covering multiple fish species, growth stages, farming modes, and temporal scales should be constructed, and robust, transferable, and low annotation dependent analytical methods should be developed. On the other hand, multimodal fusion, lightweight models, edge computing, and intelligent control technologies should be more closely integrated to promote deep coupling between feeding behavior analysis and growth models, environmental feedback, feeding devices, and aquaculture management systems. By achieving a dynamic balance among feeding efficiency, environmental stability, fish health, and economic benefit, fish feeding behavior analysis is expected to become an important enabling technology for smart aquaculture and sustainable aquaculture management.

## CRediT authorship contribution statement

Shulong Zhang: Conceptualization, Investigation, Visualization, Writing–original draft, Writing–review & editing. Daoliang Li: Writing–review & editing, Methodology. Jiayin Zhao: Writing–review & editing, Visualization. Mingyuan Yao: Writing–review & editing, Methodology. Yingyi Chen: Writing–review & editing, Methodology. Haihua Wang: Methodology, Funding acquisition, Supervision, Writing–review & editing.

## Acknowledgments

This work was supported by National Key Research and Development Program of China (2023YFD2400400, 2023YFD2400401), R&D of Key Technologies and Equipment for Aquaponics Intelligent Factory (CSTB2022TIAD-ZXX0053) and Mandarin Fish Factory Farming Service Project (202305510811525).

## Conflict of interest

The authors declare no conflicts of interest.

## Data Availability Statement

Data sharing not applicable to this article as no datasets were generated or analyzed during the current study.